\RequirePackage{fix-cm}
\documentclass[smallextended]{svjour3}

\smartqed

\usepackage{hyperref}
\usepackage{graphicx}
\usepackage{color}

\def\makeheadbox{\relax}

\begin{document}

\title{Towards Computational Awareness in Autonomous Robots:\\An Empirical Study of Computational Kernels
\author{Ashrarul H. Sifat \and Burhanuddin Bharmal \and Haibo Zeng \and Jia-Bin Huang \and Changhee Jung \and Ryan K. Williams
}

\institute{Ashrarul H. Sifat, Burhanuddin Bharmal, Haibo Zeng and Ryan K. Williams \at
              Department of Electrical and Computer Engineering, Virginia Tech, Blacksburg, VA 24061, USA \\
              \email{ashrar7@vt.edu}         \\
           Jia-Bin Huang \at Department of Computer Science, University of Maryland, College Park, MD 20742, USA
           \and
           Changhee Jung \at Department of Computer Science, Purdue University, West Lafayette, IN 47907, USA}}

\date{Received: November 17, 2021 / Accepted: TBD}

\maketitle

\begin{abstract}
The potential impact of autonomous robots on everyday life is evident in emerging applications such as precision agriculture, search and rescue, and infrastructure inspection. However, such applications necessitate operation in unknown and unstructured environments with a broad and sophisticated set of objectives, all under strict computation and power limitations. We therefore argue that the computational kernels enabling robotic autonomy must be \emph{scheduled} and \emph{optimized} to guarantee timely and correct behavior, while allowing for reconfiguration of scheduling parameters at runtime. In this paper, we consider a necessary first step towards this goal of \emph{computational awareness} in autonomous robots: an empirical study of a base set of computational kernels from the resource management perspective.  Specifically, we conduct a data-driven study of the timing, power, and memory performance of kernels for localization and mapping, path planning, task allocation, depth estimation, and optical flow, across three embedded computing platforms.  We profile and analyze these kernels to provide insight into scheduling and dynamic resource management for computation-aware autonomous robots.  Notably, our results show that there is a correlation of kernel performance with a robot's operational environment, justifying the notion of computation-aware robots and why our work is a crucial step towards this goal.
\end{abstract}

\section{Introduction}
\label{intro}
\begin{figure*}[tb]
\centering
\includegraphics[width=\textwidth]{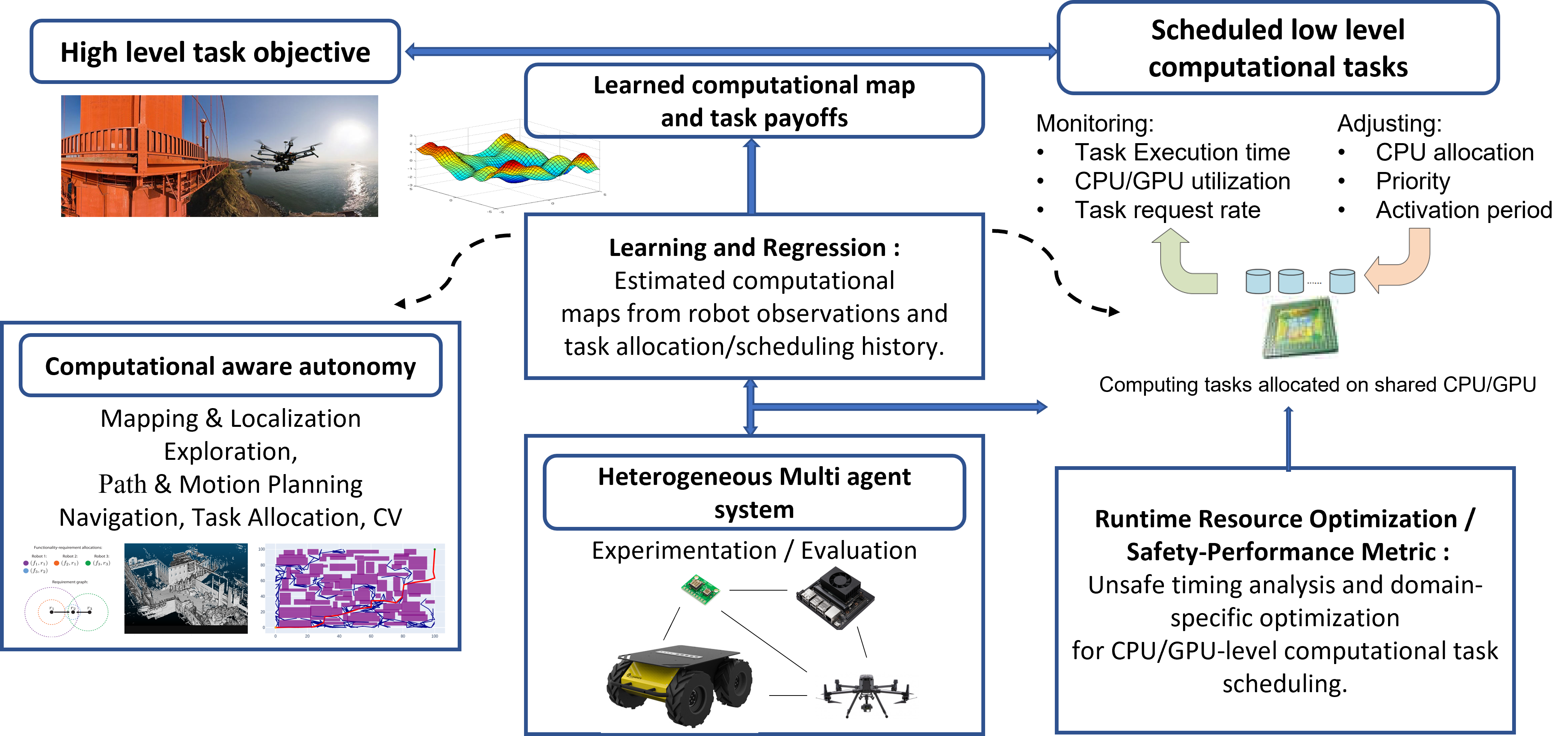}
\caption{Framework for computationally aware autonomy for timely and resilient systems.}
 \label{fig:framework}
\end{figure*}

Robots are demonstrating the ability to tackle real-world problems \emph{if} the problems are well-defined and composed of a reasonably small, fixed set of high-level tasks, e.g., welding on an assembly line \cite{Elmorov2020}.  However, our understanding remains limited for \emph{autonomous robots} that solve complex problems composed of a wide range of difficult computational tasks.  Consider for instance a search and rescue problem where autonomous robots must actively collaborate with human searchers in complex, uncertain environments (e.g., wilderness or disaster sites) \cite{Williams2020-sv,Heintzman2021-mw}.  In such a problem, an autonomous robot will be expected to execute (at least) the following \emph{base set of computational kernels}: simultaneous mapping and localization~\cite{4084563}; path planning, navigation and control~\cite{nav2010,Heintzman2021-qr,Heintzman2020-cz,Williams2017-qq,Mukherjee2017-ml,Williams:2013bh,santilli:2022}; high-level decision-making such as task allocation~\cite{8746239,Liu2021-gw,Sung2020-tl,liu2020monitoring}; and a range of computer vision (CV) tasks such as optical flow~\cite{sun2018} and depth estimation~\cite{Ranftl2019}. Furthermore, the remoteness of the operational environment \emph{necessitates on-board computation by the robots}, placing strict limits on computation and power.  Finally, as the robots work in concert with humans, there must exist \emph{guarantees} on robot behavior, and thus computation, to ensure safety.  We argue that the search and rescue example motivates the need for robots that are \emph{computation-aware}, allowing the computational kernels that support autonomy to be scheduled and optimized on embedded hardware in a manner that adapts to changing objectives and uncertain environments, while guaranteeing the timeliness and correctness of autonomous behavior. The holistic overview of such a system, depicted in Figure \ref{fig:framework} is proposed.

\begin{figure*}[tb]
\centering
\includegraphics[width=\textwidth]{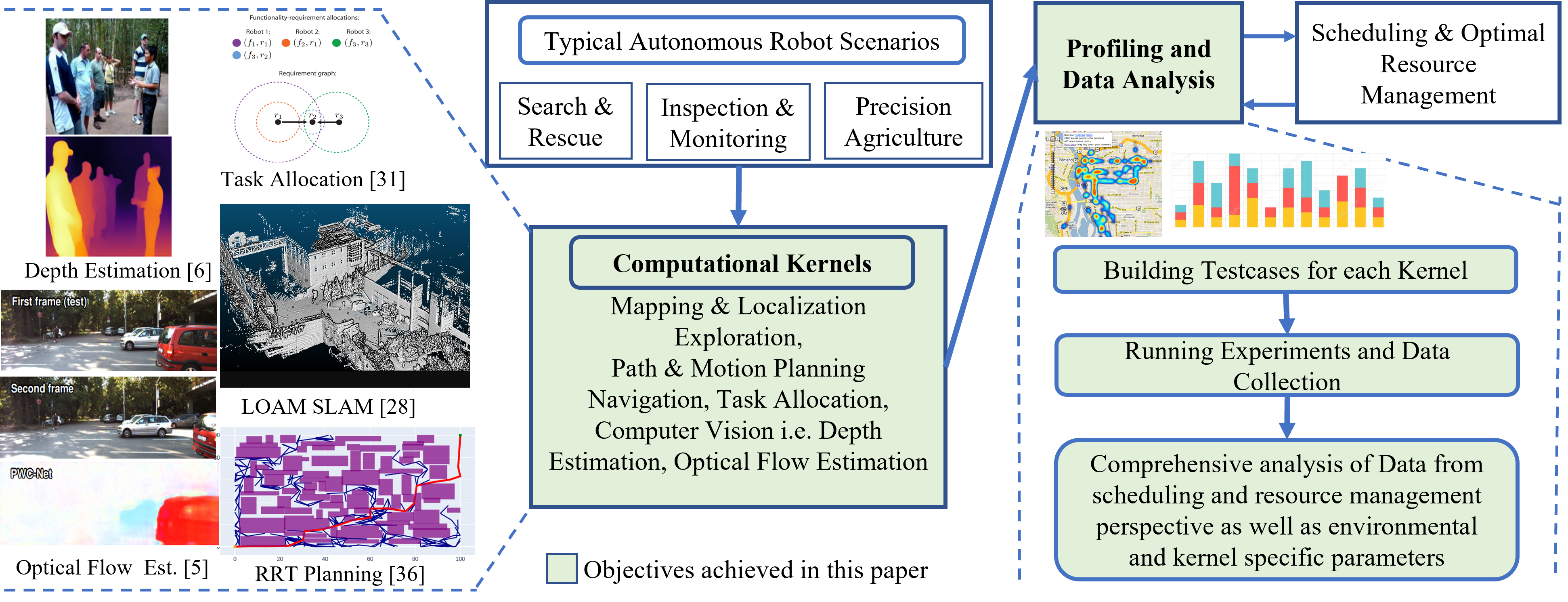}
\caption{The list of representative computational kernels and the proposed workflow for achieving computational awareness in autonomous robots with objectives achieved in this paper.}
 \label{fig:overview}
\end{figure*}

Towards this vision of \emph{computational awareness in autonomous robots}, we must first understand the computational requirements for the base set of kernels that support robotic autonomy (listed above).  In the current literature, there is a deep understanding of the aforementioned computational kernels from a \emph{theoretical} point of view, for example yielding insights into worst-case computational complexity~\cite{Guclu2018,Belavadi2017FrontierET,norrene}.  However, given a particular computational platform (such as the NVIDIA Xavier~\cite{Xavier} studied in this work), there is very little existing work on the relationship between the high-level tasks an autonomous robot performs and the underlying computational requirements for performing such tasks in a timely manner. Additionally, there have been recent efforts to integrate both hard and soft real-time scheduling-capable kernels with the Robotic Operating System (ROS) \cite{WEI2016171}, \cite{DELGADO20198}, which can form the basis for guaranteeing the timeliness and correctness of autonomous behaviors, and with ROS 2.0, real-time scheduling support specialized for robotic applications is in sight \cite{10.1145/2968478.2968502}. It is therefore critical that we build an understanding of computational kernels for autonomous robots so that such capabilities can be fully exploited in the near future.  Thus, this paper performs an empirical study of the computational requirements of a base set of kernels depicted in Figure~\ref{fig:overview}, which is the first step for the \emph{learning and regression} in our proposed workflow for achieving computational awareness in autonomous robots.

\noindent
\textbf{Related Work and Contribution.} While there is very little existing work on computational awareness in autonomous robots, we can point to a few recent efforts that are related to this paper.  A recent work focuses on benchmarking connected and autonomous vehicles (CAVs) applications from the quality of service (QoS) perspective~\cite{8567655}. The benchmarking done in this work is solely concentrated on CAVs and derive insights such as the need for heterogeneous architecture, high memory bandwidth and optimized cache architecture, which we also recognise in our systems. However, it lacks the in-depth analysis of each computational kernel that might need to run in parallel in a mobile robot unlike the CAVs and can have a different wide range of parameters to consider. Similarly, a specialized benchmark for mobile vision algorithms is presented in~\cite{6114206}, which provides a useful micro-architectural analysis of vision kernels. However, it also lacks the insight into the bigger spectrum of robotic computational kernels as well as environmental correlation and kernel specific parametric study. Kato et al.~\cite{Kato2018} present Autoware on board, to integrate the Autoware open source software for autonomous driving~\cite{autoware} on the Nvidia Drive Px2 and subsequently measure the execution time performance. However, this work lacks sufficient understanding on how the computation correlates with the driving environment.  Zhao et al.~\cite{zhao2019} perform a field study for autonomous vehicles (AV) to gain insights on how the computing systems should be designed. This work looks into the sequential operation of the AV computational kernels from a safety perspective and present a innovative safety criteria for AV. Both of these works \cite{Kato2018,zhao2019} are solely concentrated on AV which have different constraints than mobile robots such as an Unmanned Aerial Vehicle (UAV). Finally, Carlone and Karaman~\cite{carlone2018} consider the visual-inertial navigation problem, and propose anticipation with environmental cues to build a simplified model on robot dynamics and produce a computationally efficient system. However, it is an analysis of just one kernel, naturally not enough to implement a general framework such as our proposed solution. All in all these literature give valuable insight into need for computational awareness and specific studies but not the comprehensive analysis for autonomous robotic platforms.

\section{PRELIMINARIES}
Autonomous robots continue to suffer from strict limits on computation and power, especially for small aerial vehicles.  From a resource management perspective, we therefore focus on profiling that can inform two primary techniques:  (1) thread-level scheduling that prioritizes among multiple computational kernels competing for shared computational resources such as CPU, GPU, and memory \cite{Linus2020,Fang2020,6237036,6714422}; and (2) the Dynamic Voltage and Frequency Scaling (DVFS) mechanism on modern processors that allows for fine-grained frequency scaling and power management \cite{6986119}.   We aim to analyze each computational kernel for sustainability and predictability by showing their worst case behavior, average behavior, and the variation from them.  For an autonomous robot, we argue that a static, single-valued worst case execution time (WCET) is too pessimistic as the robot's computational needs are highly dynamic and potentially dependent on the environment and goal.  In order to take advantage of the above techniques in a robotics setting, we profile the utilization, timing, memory, and power consumption properties of a base set of computational kernels that support autonomy.

\begin{figure}
 \centerline{\includegraphics[width=4.0in]{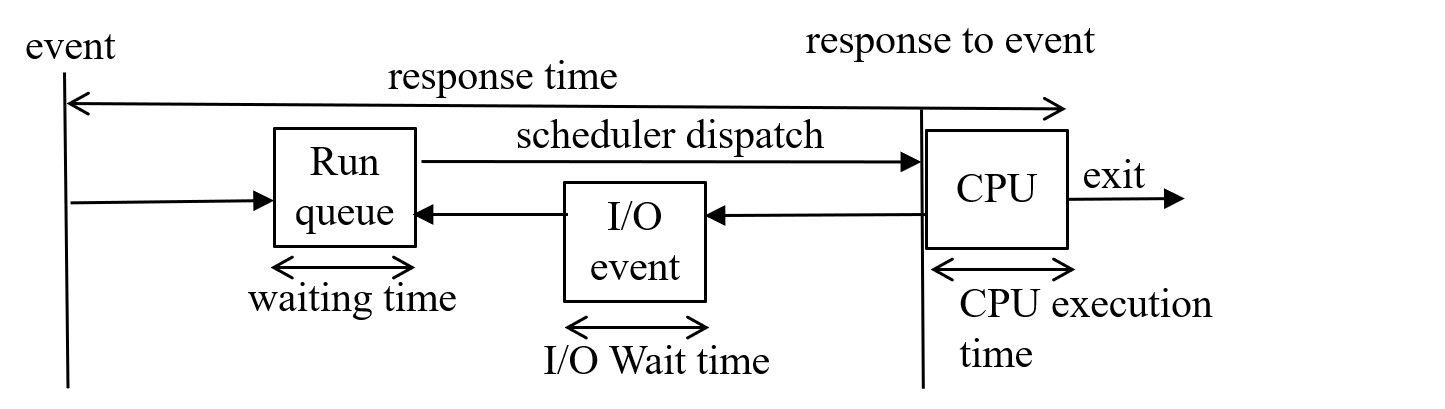}}
 \caption{CPU scheduling parameters and operation.}
 \label{sched}
\end{figure}

There are several different parameters to consider when selecting the most appropriate scheduling algorithm. Figure \ref{sched} depicts the general operation of a scheduler, with definitions for the parameters we aim to measure given below.
\begin{itemize}
    \item \textit{CPU Utilization}: The ratio of a CPU's load and the computational capacity of the CPU.
    \item \textit{Throughput}: The total number of processes completed per unit time, i.e., the total amount of work done in a unit of time.
    \item \textit{Turnaround/Response Time}: The amount of time taken to execute a particular process/thread.
    \item \textit{User Time}: The amount of time a process/thread spends in the user space of the host operating system. This is also referred to as CPU execution time.
   \item \textit{System Time}: The amount of time a process spends in kernel level instructions such as interrupts, I/O calls, etc.
   \item \textit{Activation Period}: The interval  or period after which each periodic kernel is repeated and therefore dispatched to the CPU scheduler.
\end{itemize}
In general, CPU utilization and throughput are maximized and other factors are minimized to achieve performance optimization. While the response time is varied widely across different systems and scheduling policies, the execution time is tied most closely to the specifics of hardware and the computational kernel itself. Therefore, in this paper we focus our profiling on the execution time in user space and kernel space as a means of characterizing the requirements for our computational kernels.  At the same time, thread-level execution time breakdown and CPU-affinity are also a necessary part of scheduling. Therefore, among the other parameters of analysis we will provide the number of processes per computational kernel, number of threads per process, average utilization of threads, memory utilization, and power consumption. 

We also investigate relevant parameters for GPU-intensive vision kernels, including instructions per cycle (IPC), shared utilization, various (xyz) functional unit utilization, achieved occupancy, and various memory load throughput. As the first step towards optimizing and guaranteeing the timeliness of computation for an autonomous robot, we believe these parameters are of vital importance to understand. GPU performance parameters also vary significantly across different architectures. The definitions of the parameters we measure are given below:
\begin{itemize}
    \item \textit{Achieved Occupancy}: Ratio of the average active warps per active cycle to the maximum number of warps supported on a GPU.
    \item \textit{xyz FU Utilization}:	The utilization level of the GPU xyz functional units on a scale of 0 to 10. Here xyz depends on the hardware platforms. Typically xyz = alu, cf, special etc.
    \item \textit{Warp Execution Efficiency}:  Ratio of the average active threads per warp to the maximum number of threads per warp supported on a GPU.
    \item \textit{Shared Utilization}:  The utilization level of shared memory relative to peak utilization on a scale of 0 to 10.
    \item \textit{GPU Utilization}:  Percent of time over the past sample period during which one or more kernels was executing on the GPU. (Only for Zotac platform using \emph{Nvidia-smi} interface)  
    \item \textit{GPU Memory Usage}:  Total memory allocated by active contexts. (Only for Zotac platform using \emph{Nvidia-smi} interface). The GPU memory utilization can be derived by scaling this value (\%) to the GPU memory capacity.  
\end{itemize}

Finally, it is worth noting that the computational complexity of the tasks carried out by an autonomous robot can be used to \emph{mathematically} formulate the execution time required for each CPU process.  However, the complex nature of CPU execution cycles combined with the hardware and kernel related variables and dependencies makes it infeasible to predict computational needs merely from a mathematical formulation. Hence, we begin in this paper with profiling of our computational kernels in various settings across popular computational platforms for robotics. Only then is it possible to answer the pressing questions of combining computational kernels, implementing appropriate scheduling policies, and determining the rates at which we can execute the kernels.

\section{Computational Kernels}
This section provides a brief overview of the concept and operations of the computational kernels we profile. Note that \textit{each computational kernel may be broken into multiple threads or even processes}, as detailed in Section~\ref{sec:empirical}. 

\subsection{Simultaneous Localization and Mapping}

Simultaneous Localization and Mapping (SLAM) is an age-old problem in robotics which still attracts significant interest from the robotics community. Indeed, SLAM enables a robot to build an understanding of its environment over time and is thus foundational to robotics applications (see the left hand side of Figure~\ref{fig:overview}  for a depiction of typical SLAM output).  In this work, we select two point cloud-based methods for SLAM to study:  Gmapping \cite{KIM201838} and Lidar Odometry and Mapping (LOAM) \cite{loam}, which are commonly applied in robotics.  We select two versions of SLAM to demonstrate two related concepts:  (1) computational kernels for a given task can vary significantly in performance; and (2) different kernel versions may be used interchangeably based on computational/power limitations and robot objectives.

\begin{figure}
 \centerline{\includegraphics[width=4.0in]{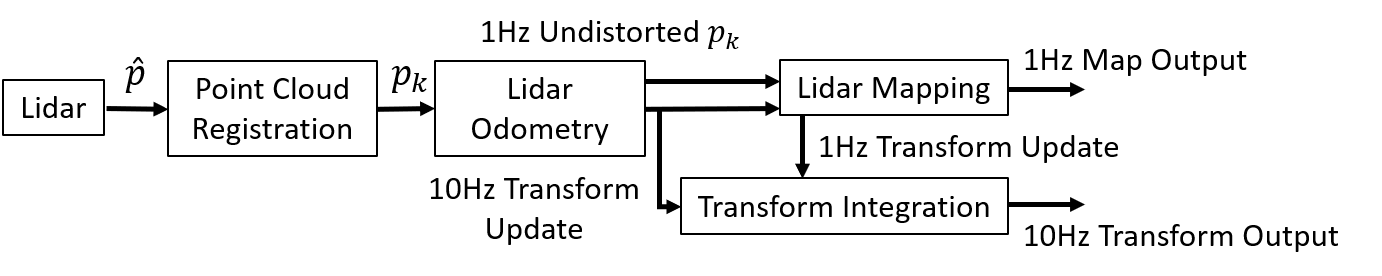}}
 \caption{Block Diagram of Loam Software System \cite{loam}.}
 \label{loam}
\end{figure}

As one of the most popular computational kernels for outdoor SLAM, Gmapping consists of the Rao-Blackwellized particle filer to learn grid maps from laser range data. This approach uses a particle filter in which each particle carries an individual map of the environment.  It is one of the pioneering approaches to SLAM~\cite{4084563}, with a tractable computational complexity of $O(NM)$, where $N$ is the number of particles and $M$ is the size of the corresponding grid map.  We also profile LOAM which produces real-time 3D maps from Lidar point cloud data, which are generally of a higher quality than Gmapping. LOAM has four parallel process nodes running simultaneously to update the mapping and odometry data and produces updates on maps in real-time (see Figure \ref{loam}). 

\begin{remark}
It is important to note that the generation of maps and localization solutions is highly dependent on the features of the environment.  We will demonstrate that this phenomenon can be seen directly in the computational profiling for SLAM, where we contrast between indoor and outdoor environments in terms of execution times and CPU usage.
\end{remark}

\subsection{Path Planning}

Path planning is another critical computational kernel for autonomous robots (see Figure~\ref{fig:overview} left for a depiction of typical planner output).  In this paper, we focus on planning methods based on Rapidly-Exploring Random Trees (RRTs) \cite{844730}, as they are very popular and there are numerous RRT variants for various application scenarios. We profile a 2D RRT method which utilizes R-tree \cite{RTREE16} to improve performance by avoiding point-wise collision and distance checking.  When profiling the RRT kernel, we generate an environment with random obstacles to represent dynamic and unknown environments that the robot is presumed to traverse. Importantly, the execution time of RRT will depend on the size of the environment, the number of planning iterations allowed, the goal location to reach, obstacles, and the step or sampling size of the planner. In this work, we investigate the effect of step/sampling size (R), environment size, and obstacle density on RRT computational requirements.

\subsection{Task Allocation}
Single robots and multi-robot teams often have multiple high-level tasks to complete over a mission, with the tasks potentially streaming in or changing in nature over time (see the left hand side of Figure~\ref{fig:overview}  for a depiction of task allocation).  To model the autonomous decision-making necessary to allocate robot resources to tasks optimally, we profile task allocation problems which are difficult combinatorial optimization problems with resource constraints. In this paper, we benchmark and analyze a novel task allocation problem that aims to allocate a set of robots' functionalities to a mission such that certain functionality combinations are satisfied.  A combinatorial optimization problem with matroid constraints is solved using a greedy algorithm, yielding a solution with bounded suboptimality~\cite{Williams2017,8746239}. The parameters of interest for this task allocation kernel are the robot cardinality, $N$, requirement cardinality $R$, and functionality cardinality $F$.  By changing these parameters, we can alter the complexity of the task allocation problem and profile its computational requirements.
\subsection{Computer Vision Kernels}
To support a robot navigating an environment safely while avoiding \emph{dynamically moving objects} (such as pedestrians), we study two additional kernels for computer vision tasks (see Figure~\ref{fig:overview} left for a depiction of the typical input/output). Specifically, in this work, we focus on 1) optical flow (for estimating dense motion between consecutive frames) and 2) monocular depth estimation (for predicting dense scene depth). In each kernel, we choose the state-of-the-art models of PWC-Net~\cite{sun2018} for optical flow and MiDaSv2~\cite{Ranftl2019} for depth estimation. For both models, we use the publicly available pre-trained models in PyTorch for our empirical analysis.

\section{Empirical Analysis} \label{sec:empirical}
\subsection{Mobile Robot Platform}
\begin{figure}
 \centerline{\includegraphics[width=1.8in]{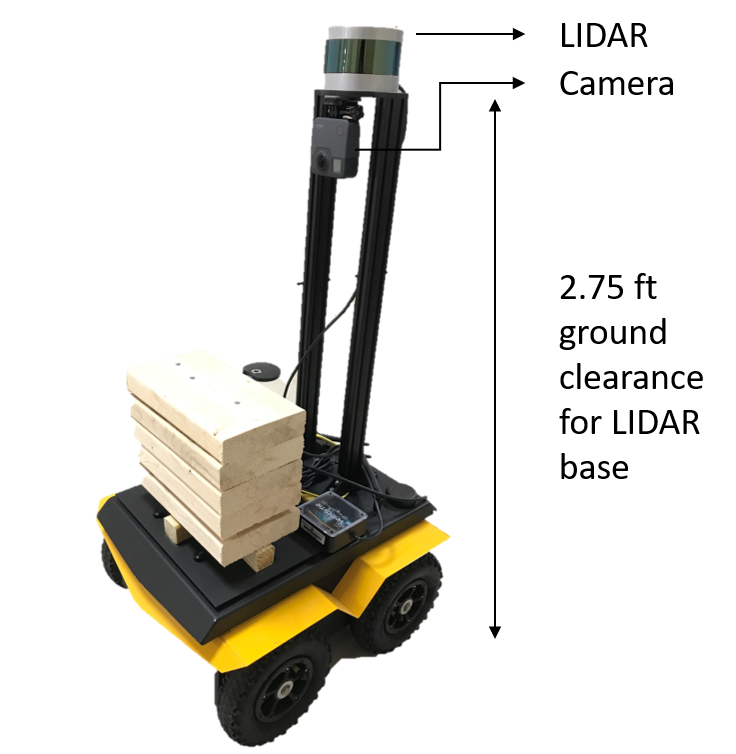}}
 \caption{Platform for data collection. Velodyne LIDAR and Go-Pro Fusion 360 camera are mounted on a Jackal UGV. }
 \label{robot}
\end{figure}

As our motivation in this work is to deploy computation-aware robots to solve real-world problems, we aimed to exploit real-world data for profiling where possible. Sensor data collection for several computational kernels was conducted using the Jackal Unmanned Ground Vehicle (UGV) platform from Clearpath robotics \cite{jackal}. The UGV was equipped with the Velodyne VLP-16 LIDAR and a Go-Pro Fusion 360 camera, with the sensors mounted as seen in Figure \ref{robot}. 
\subsection{Computational Platforms}
We consider three representative hardware platforms for robots. Their technical specifications are summarized in Table~\ref{Platform_compute}. Our first platform is the NVIDIA Jetson AGX Xavier \cite{Xavier} which has significant computational capacity for autonomous robots in a compact form factor.  Additionally, the Xavier has a heterogeneous architecture based on a System-on-Chip (SoC) that supports both CPU and GPU computation. The second computational platform we consider is the NVIDIA Jetson Xavier NX \cite{jxnx} which is the state-of-the-art embedded platform from NVIDIA that is suited for drones and smaller robots with strict resource constraints. Both Xavier devices have shared memory between the CPU and GPU which along with their architecture makes them extremely power efficient. Finally, we profile our kernels on the Zotac Mini PC system with specifications similar to a desktop computer, but with a form factor that allows integration into a larger robot with high power availability. 

From the software perspective, the Linux environment is the most popular open source operating system used for robotics worldwide. Linux coupled with ROS is the go-to system for robotics researchers. Thus, we profile our platforms/kernels in Ubuntu 18.04 with ROS middleware (ROS Melodic). The kernel version of Linux for the Xavier platforms is 4.9.140-tegra which is the latest ARM-compatible version of the vanilla Linux kernel. For the Zotac platform, we use the Linux kernel 5.4.

\begin{table}
\caption{Computational Platforms specifications.}
\label{Platform_compute}       
\resizebox{\columnwidth}{!}{%
\begin{tabular}{llllllllll}
\hline\noalign{\smallskip}
  Device & \multicolumn{4}{c}{CPU}  & \multicolumn{2}{c}{GPU}  & Memory & Power Range & Dimension \\ \cline{2-7}
  & Architecture & cores & Maximum Frequency & cache & Architecture & Cores & & & (mm)\\
\noalign{\smallskip}\hline\noalign{\smallskip}
 Nvidia Jetson   & ARM v8.2 64-bit & 8 & 2.26GHz & 8MB L2  & Volta & 512 & 16GB & 10W - 72W & 105$\times$105$\times$65\\
      Xavier AGX & & & & + 4MB L3 & (Integrated) & &  & &\\
\noalign{\smallskip}\hline\noalign{\smallskip}
  Nvidia Jetson & Nvidia Carmel  & 6 & 2.26GHz &  6MB L2 & Volta & 384 & 8GB & 10W - 15W & 103$\times$90.5$\times$34.66 \\
    Xavier NX  & ARM v8.2 64-bit & & &+ 4MB L3 & (Integrated) & &  & & \\
  \noalign{\smallskip}\hline\noalign{\smallskip}  
     Zotac  & Intel Core  & 6$\times$2 & 4.5Ghz & 256KB L2 & Turing &  2304 & 32GB & 330W &210$\times$203$\times$62.2 \\
  EN72070V &i7-9750H & & & + 12MB L3 & (Discrete, 8GB)& &  & &\\
      
\noalign{\smallskip}\hline
\end{tabular}}
\end{table}

\subsection{Profiling Data Acquisition}
For CPU execution time, utilization, and shared memory usage, we primarily use \emph{psutil} \cite{Psutil2020}. Psutil is a python tool that implements many basic Linux command-line operations such as \emph{ps} and \emph{uptime}, among others.  However, for the task allocation kernel, CPU execution time is calculated using the \emph{times} Linux kernel api, CPU utilization is determined using \emph{mpstat}, and the \emph{vmstat} command provides memory usage.  The process and threads details for kernels were captured using NVIDIA Nsight Systems whereas GPU statistics were recorded using the NVIDIA Profiler tool \cite{nvprof} for NVIDIA Xaviers. For the Zotac platform, GPU memory usage, GPU utilization, and GPU power were recorded using the \emph{gpustat} command which is based on the \emph{Nvidia-smi} interface.

For power measurements, on-board INA power monitors in the NVIDIA platforms are utilized which allow power monitoring at runtime with an accuracy of 95\%. For the Zotac platform, we calculate CPU power usage with the \emph{energyusage} python command which calculates the power via the RAPL (Running Average Power Limit) interfaces found on Intel processors\cite{intelpowre}. 

The objective of this paper is to gain insight into the computational aspects of crucial robotic kernels. Therefore, we use these well known and rigorous tools developed over a long time and reputation, so that our results are not questionable. Developing custom tools is not in the scope of this paper, but can certainly be pursued when the optimization problem is narrowly defined such as in RESCH \cite{Kato2009}.

\subsection{Profiling Results}
In this section, we present our profiling results with only limited initial insights. The implications and usefulness of these results are discussed in detail in Section \ref{sec:disc}. As a reference, we provide a recorded video of the data collection experimentation at https://youtu.be/LvapXWqEAqU.
\subsubsection{SLAM}
\begin{figure}
 \centerline{\includegraphics[width=3in]{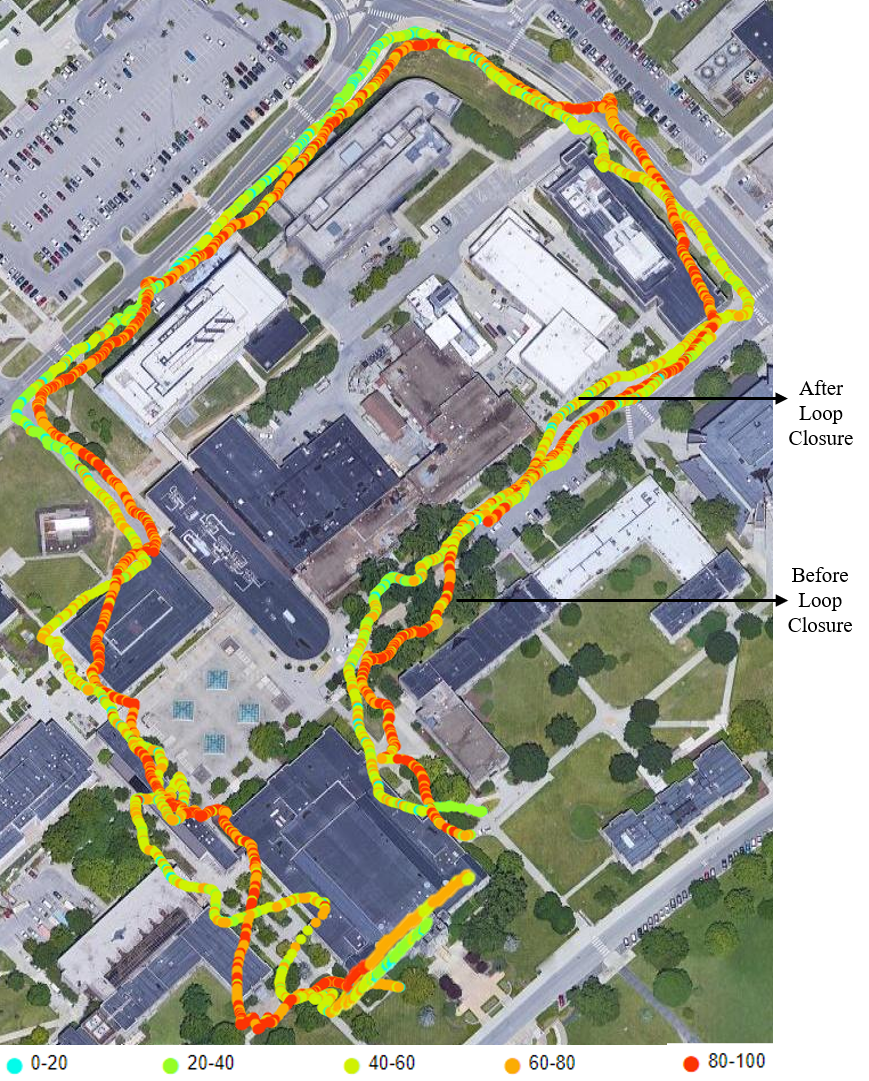}}
 \caption{Geographical locations of the data points for Figure \ref{fig:loam}. Depicted is the execution times for LOAM in the Xavier AGX platform. The colors indicate the execution time required at a particular location (red is high, green is low).}
 \label{loam_map}
\end{figure}

\begin{figure}
 \centering
  \begin{tabular}{@{}c@{}}
    \includegraphics[width=.7\linewidth]{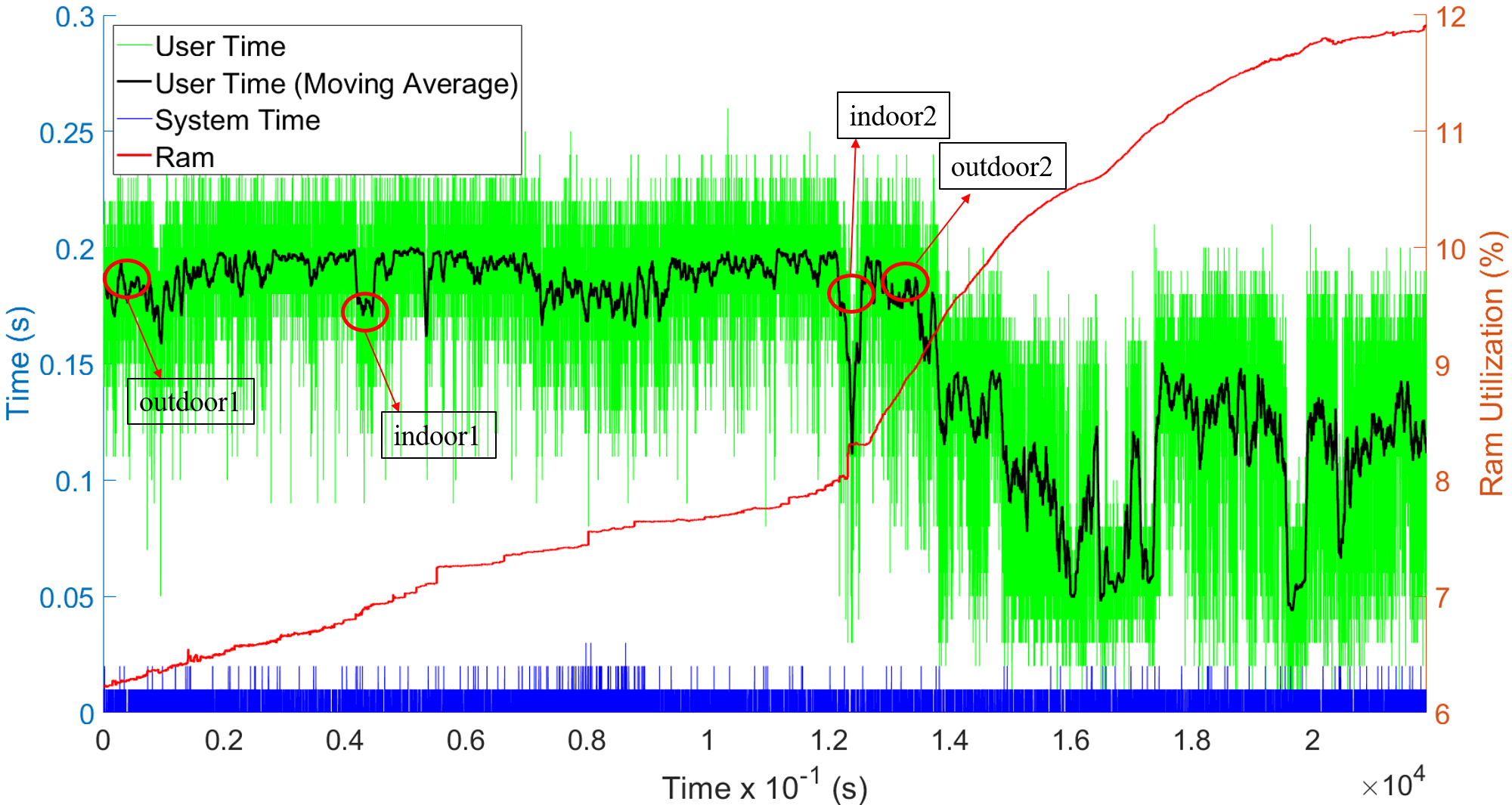} \\[\abovecaptionskip]
    \small (a)
  \end{tabular}


  \begin{tabular}{@{}c@{}}
    \includegraphics[width=.7\linewidth]{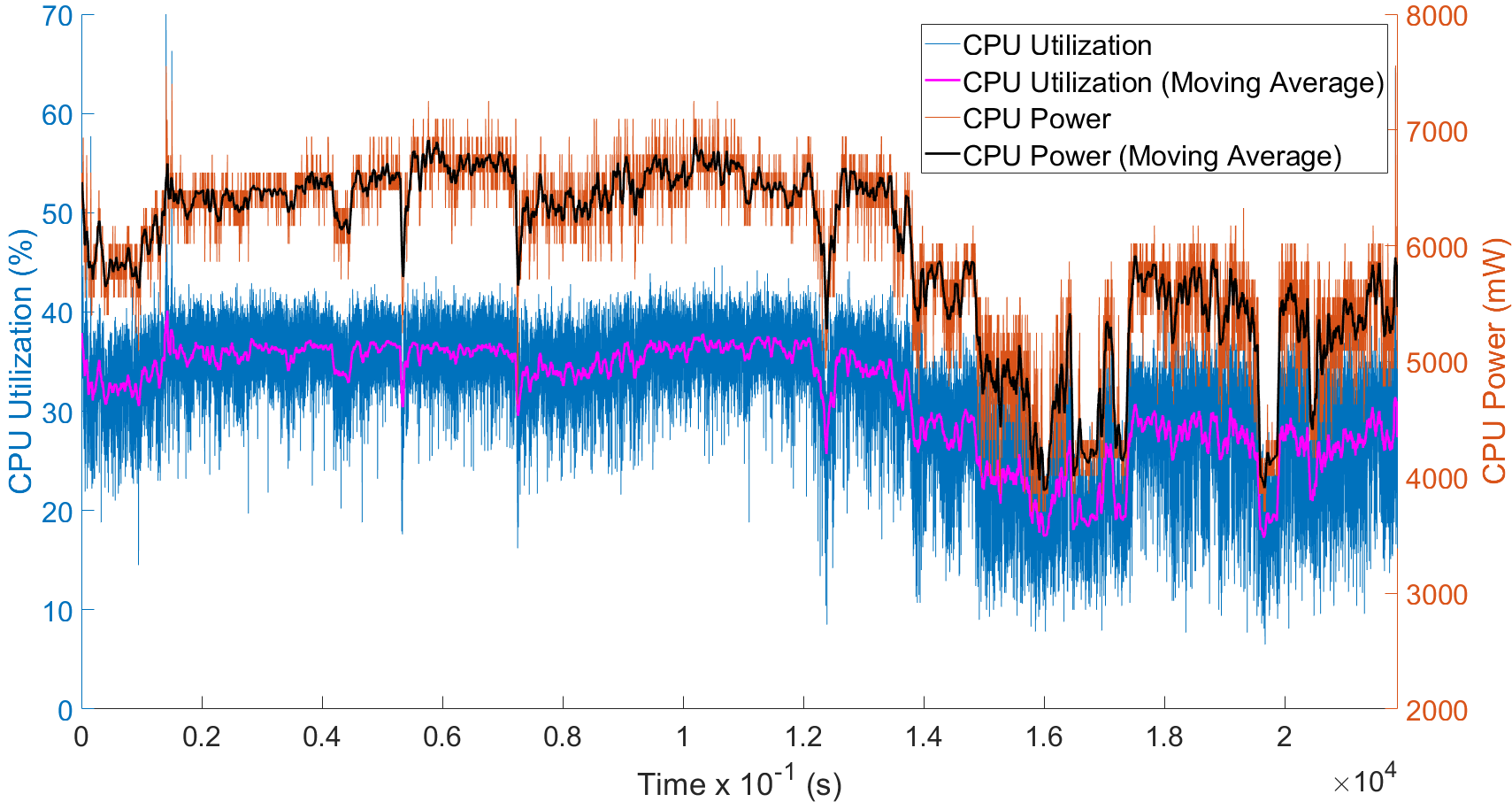} \\[\abovecaptionskip]
    \small (b)
  \end{tabular}
  

  \begin{tabular}{@{}c@{}}
    \includegraphics[width=.7\linewidth]{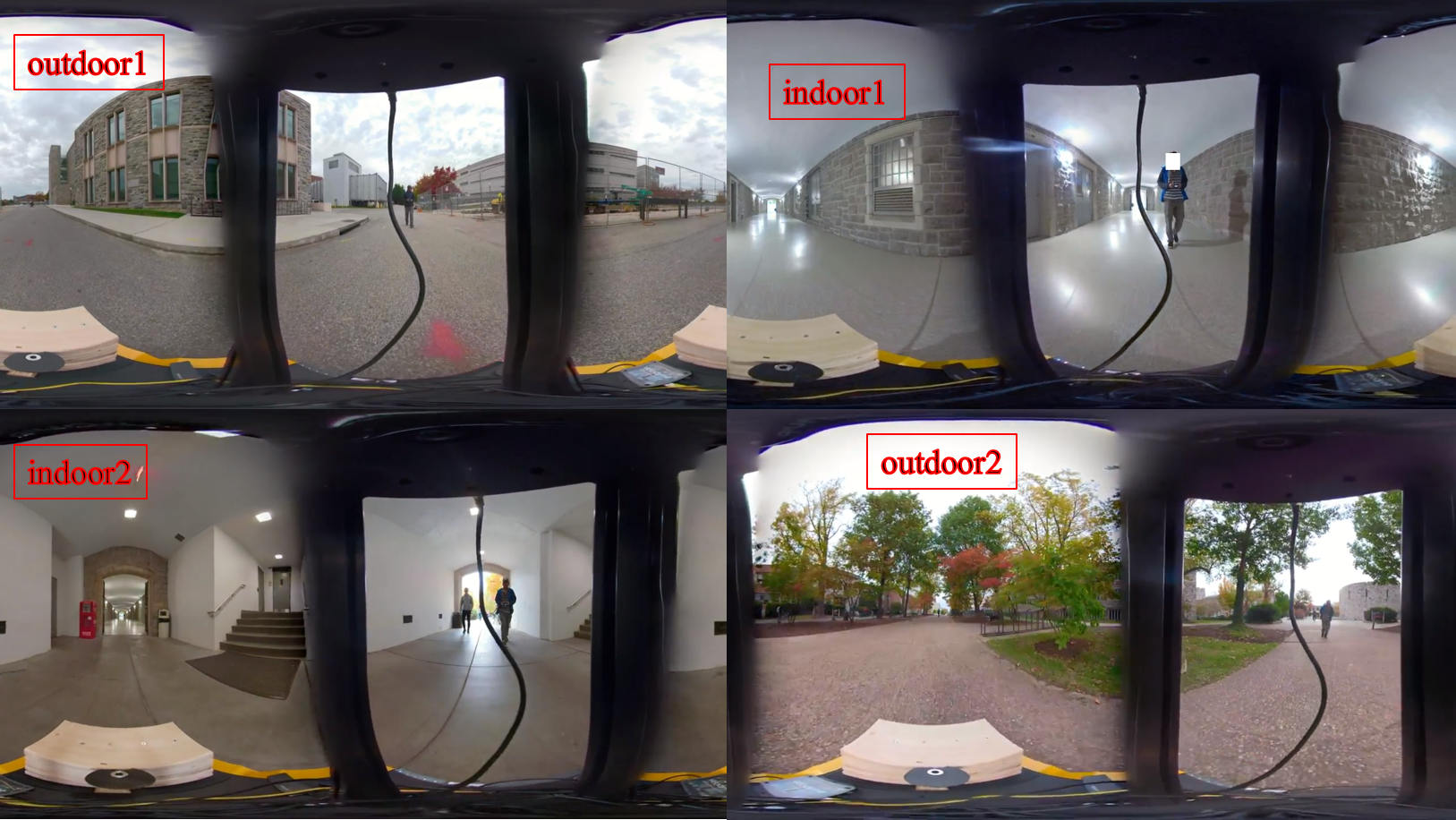}
    \\[\abovecaptionskip]
    \small (c)
  \end{tabular}

  \caption{Computational profile data of LOAM SLAM in Xavier AGX platform. (a) Execution time which includes User Time and System Times for an activation period of 0.1 sec. In the Right axis is RAM utilization as a percentage of used RAM to the total available RAM. (b) CPU Utilization and Power measurements. (c) Indoor and outdoor environment corresponding to data points circled in (a)}
  \label{fig:loam}
\end{figure}

The LOAM SLAM benchmark was carried out with the real-world data collected as described previously.  For the data presented in this paper, the same loop in a large, mostly outdoor area was traversed three times as is seen in Figure \ref{loam_map}.  During this traversal, the first completion of the loop did not achieve \emph{loop closure} in the SLAM process (feature reconciliation that improves maps as more data is collected), whereas the second and third loop completion did see loop closure. As a result, it is visible that the execution time decreases rapidly between the second and third loop which is visible in the computational profile (Figure \ref{fig:loam}).  Moreover, there was an average of \emph{30\% drop} in execution time when passing through indoor confined spaces such as a tunnel.  Finally, in Figure \ref{fig:loam}a we see that the RAM utilization increases with the traversal of the environment.  Based on the above discussion, we can deduce that repeated environment traversal, environment complexity (indoor vs.\ outdoor), and mission length correlate strongly with the computational requirements for SLAM. 

\begin{table}
\caption{Process/Thread summary for LOAM in Xavier AGX platform.}
\label{thread_loam}      
\resizebox{\columnwidth}{!}{%
\begin{tabular}{lllll}
\hline\noalign{\smallskip}
  Process & LaserMapping  & LaserOdometry & Transform  & Multi-Scan  \\
     &  &  & Maintenance & Registration \\
\noalign{\smallskip}\hline\noalign{\smallskip}
 Number of Threads & 4 & 4 & 4 & 4 \\ 
  \% Utilization of dominant thread & 98.97 & 97.72 & 60.7 & 98.07 \\
  \% of Execution Time & 32.8 & 28.8 & 0.17 & 38.13 \\
  Avg. Execution Time & 0.08448 & 0.0736 & 0.0003 & 0.0947\\ 
  Variance in Exec Time & 0.00106719 & 0.0007106 & 2.91E-6 & 0.000108 \\
\noalign{\smallskip}\hline
\end{tabular}}
\end{table}
\begin{remark}
Regarding the explanation of the memory consumption, it is highly unlikely that it has a memory leak, at least from primary inspection of the code as memory allocation was not used in it. LOAM SLAM\cite{loam} is a very well cited and high quality LIDAR based 3D SLAM technique which has high load on the processor. It has 16 threads (4x4) and 4 dominant threads from the four processes. Moreover, the scope of this paper is not to look into particular coding issues but to identify computational and environmental aspects in robotics kernels. As we do not see this attribute in the 2D SLAM discussed later, the memory buildup is likely a characteristic of 3D high definition LOAM SLAM. However, further micro-architectural analysis into this kernel would be necessary to pin down this attribute and possibly find a remedy for this.
\end{remark}

Table \ref{thread_loam} summarizes the different processes, thread counts, and percentages of execution times for the LOAM kernel in the Xavier platform. To specify more clearly, this data in this table are generated using the nvidia profiler and its definitions are a little different than the ones presented in section II. Here, The $\%$ utilization of dominant thread denotes  $\%$ utilization of a single core, the $\%$ of Execution time denotes the percentage of the particular process's execution time compared to the total execution times of these four process that make up the LOAM kernel and the average execution time is just the average of that process.
SLAM is highly dependent on sensor data and thus the ROS middleware is used to enable smooth message passing between the different SLAM processes. The Laser Mapping and Odometry processes contribute towards the map building for LOAM and thus require the most resources combined. On the other hand, the Scan Registration process only contributes towards point cloud data processing which is a constant cycle and therefore has very little variation in data. Another vital aspect is that all of these processes have one dominant thread for scheduling purposes as the other threads can be regarded as overhead and therefore makes it convenient from a scheduling perspective. From the tracing profiles in Nsight Systems, it is visible that these four processes run in parallel in different CPU cores while the ROS middleware ensures the message passing between them. Because of ROS's lack of Quality of service (QoS) in communication and real-time support, some momentary drops in packets are observed when the data of CPU performance were collected during the kernel's operation.

\begin{figure}
 \centering
  \begin{tabular}{@{}c@{}}
    \includegraphics[width=.7\linewidth]{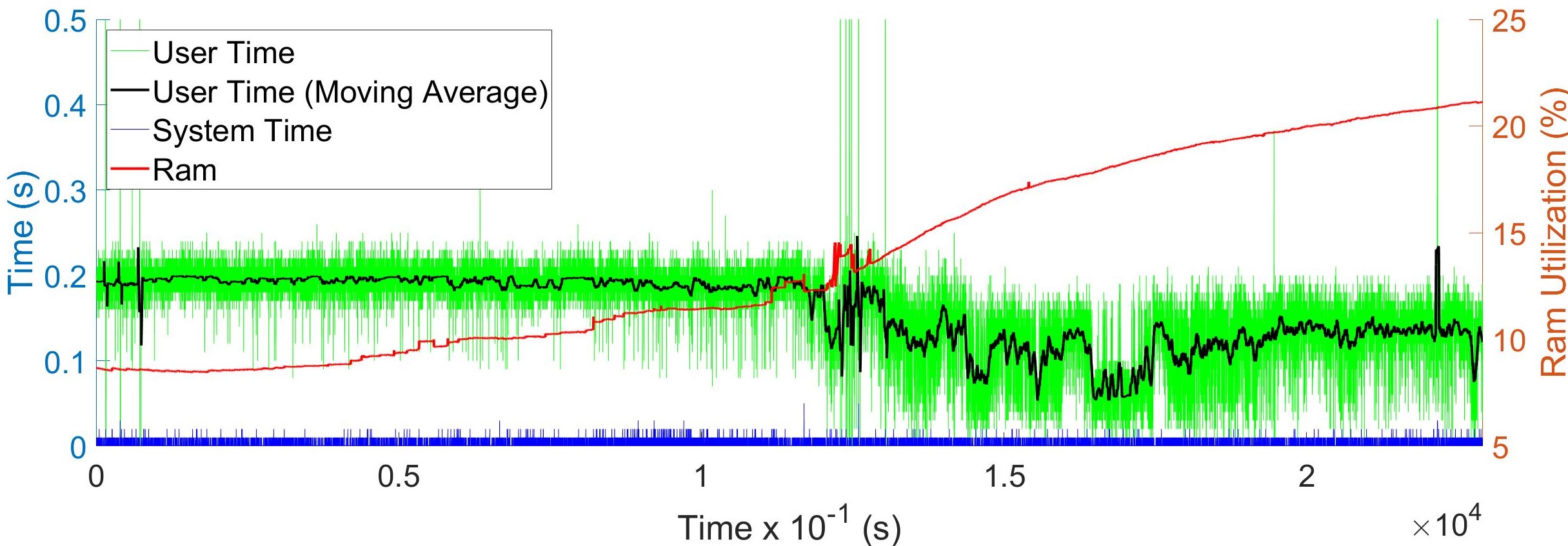} 
    \\[\abovecaptionskip]
    \small (a)
  \end{tabular}
  \begin{tabular}{@{}c@{}}
    \includegraphics[width=.7\linewidth]{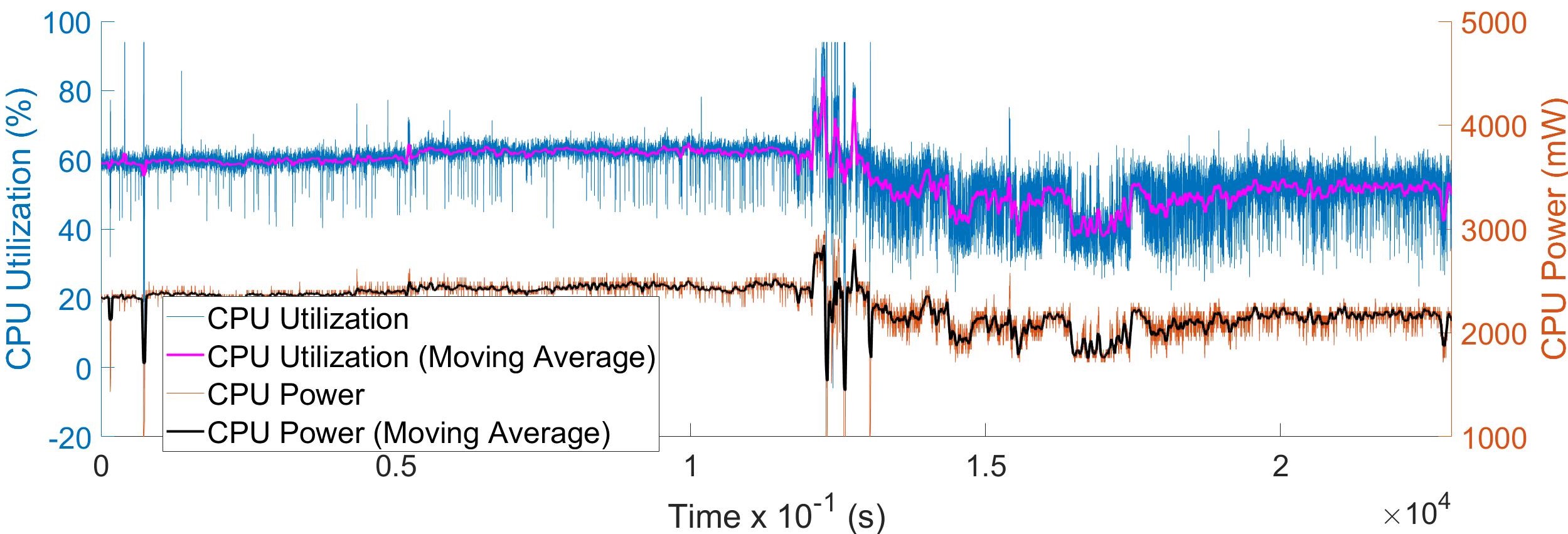} 
    \\[\abovecaptionskip]
    \small (b)
  \end{tabular}
 \caption{Computational profile data of LOAM SLAM in Xavier NX. (a) Execution time which includes User Time and System Times for an activation period of 0.1 sec. (b) CPU Utilization and Power measurements.}
  \label{fig:loamnx}
\end{figure}

\begin{figure}
 \centering
  \begin{tabular}{@{}c@{}}
    \includegraphics[width=.7\linewidth]{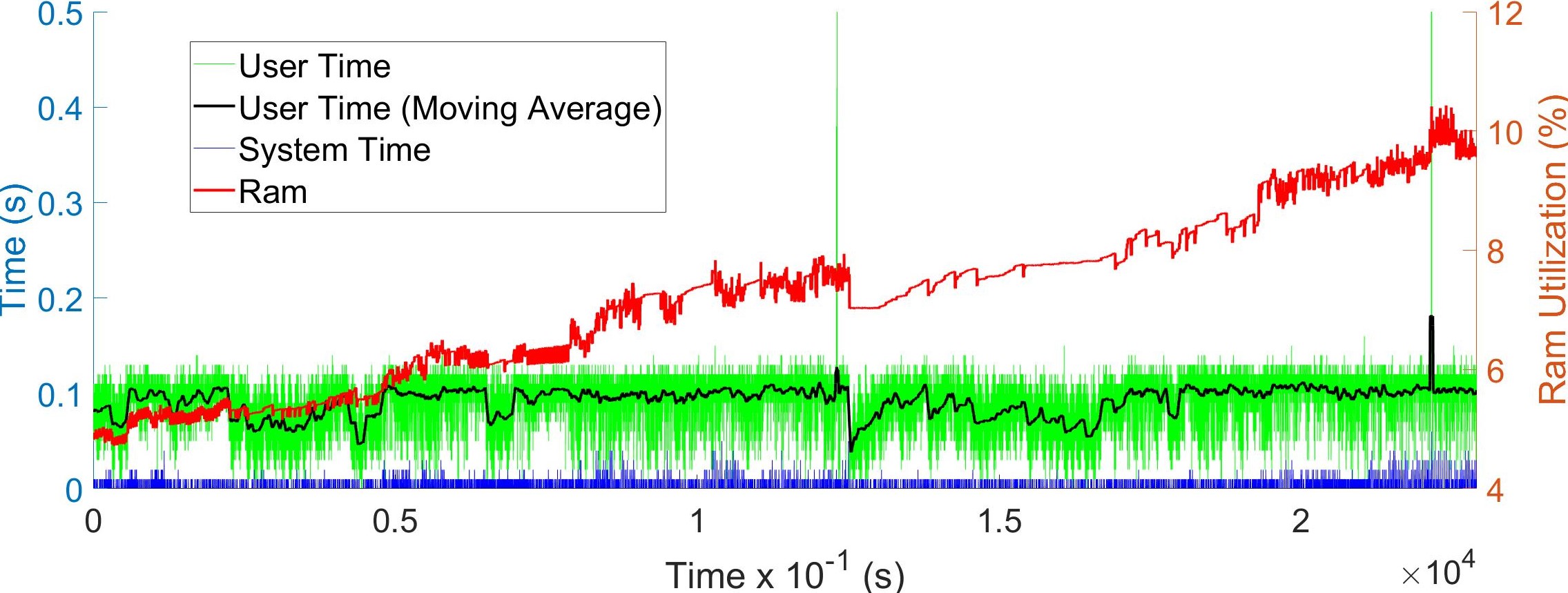} 
    \\[\abovecaptionskip]
    \small (a)
  \end{tabular}
  \begin{tabular}{@{}c@{}}
    \includegraphics[width=.7\linewidth]{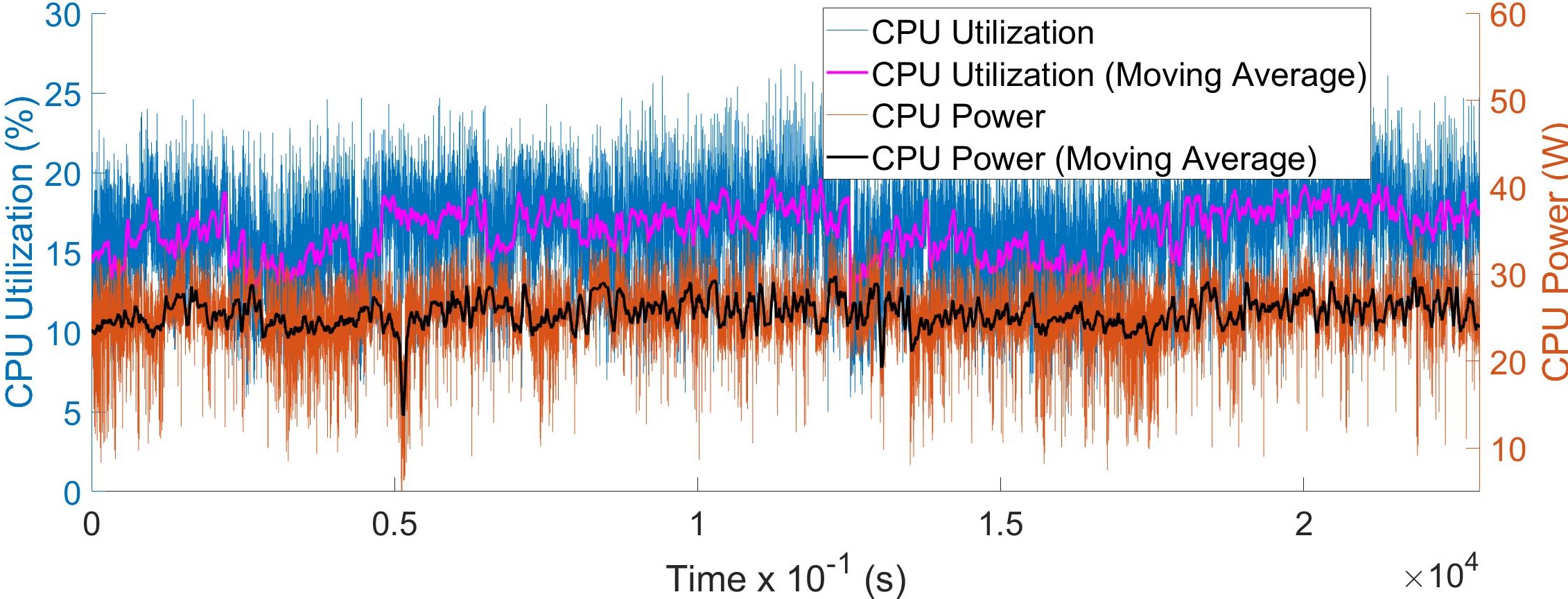} 
    \\[\abovecaptionskip]
    \small (b)
  \end{tabular}
    \caption{ Computational profile data of LOAM SLAM in Zotac. (a) Execution time which includes User Time and System Times for an activation period of 0.1 sec. (b) CPU Utilization and Power measurements.}
  \label{fig:loamnzot}
\end{figure}

Figures \ref{fig:loamnx} and \ref{fig:loamnzot} depict the LOAM performance metrics in the Xavier NX and Zotac platforms, respectively. It is visible that the execution times are mostly the same for Xavier NX but the CPU and memory utilization are significantly higher which is expected behavior. However, in the Zotac platform the execution times are significantly lower than the Xavier platforms which is also expected considering its higher frequency and higher CPU and memory capacity.

\begin{figure}
 \centering
  \begin{tabular}{@{}c@{}}
    \includegraphics[width=.7\linewidth]{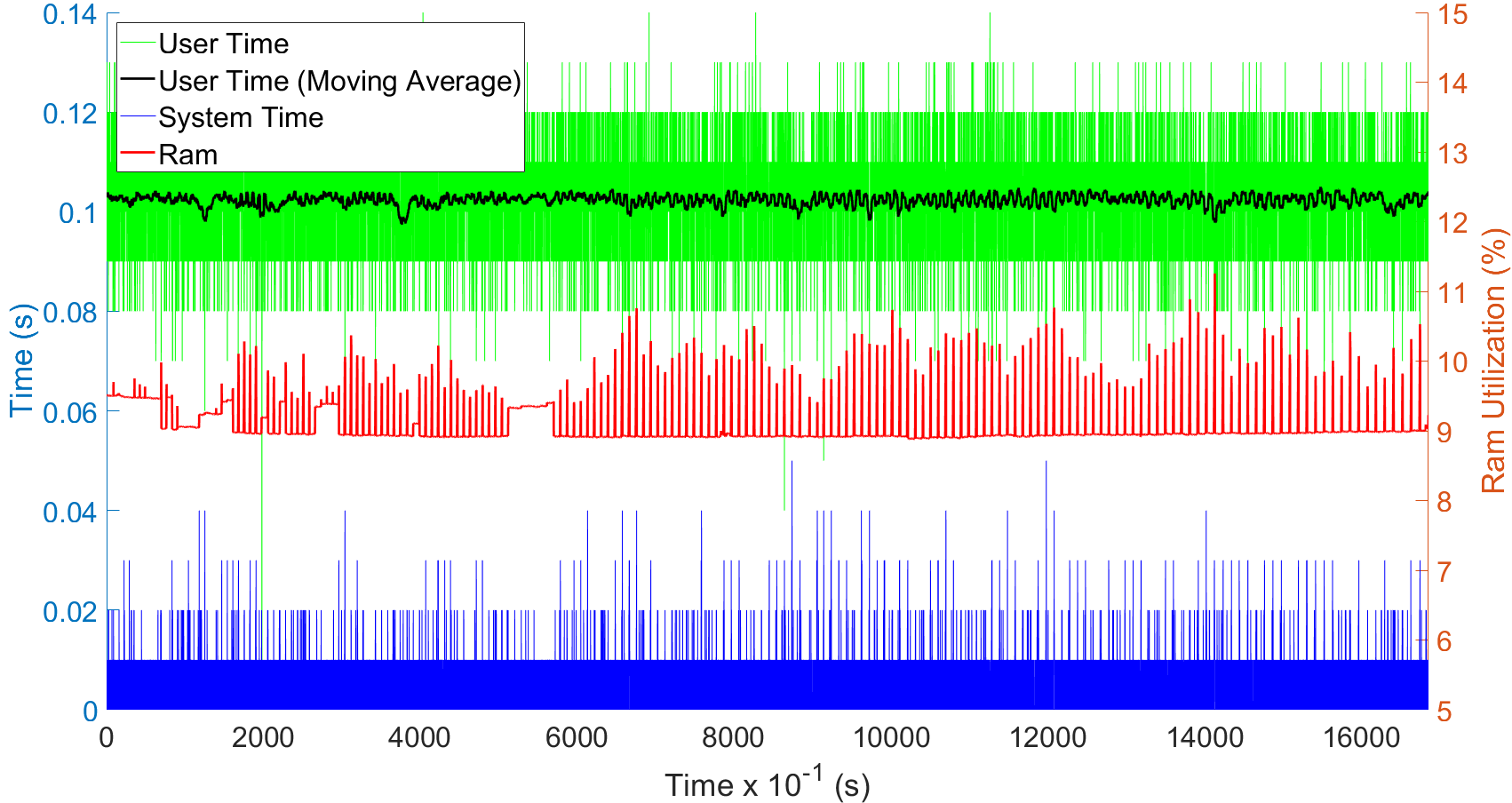} 
    \\[\abovecaptionskip]
    \small (a)
  \end{tabular}
  \begin{tabular}{@{}c@{}}
    \includegraphics[width=.7\linewidth]{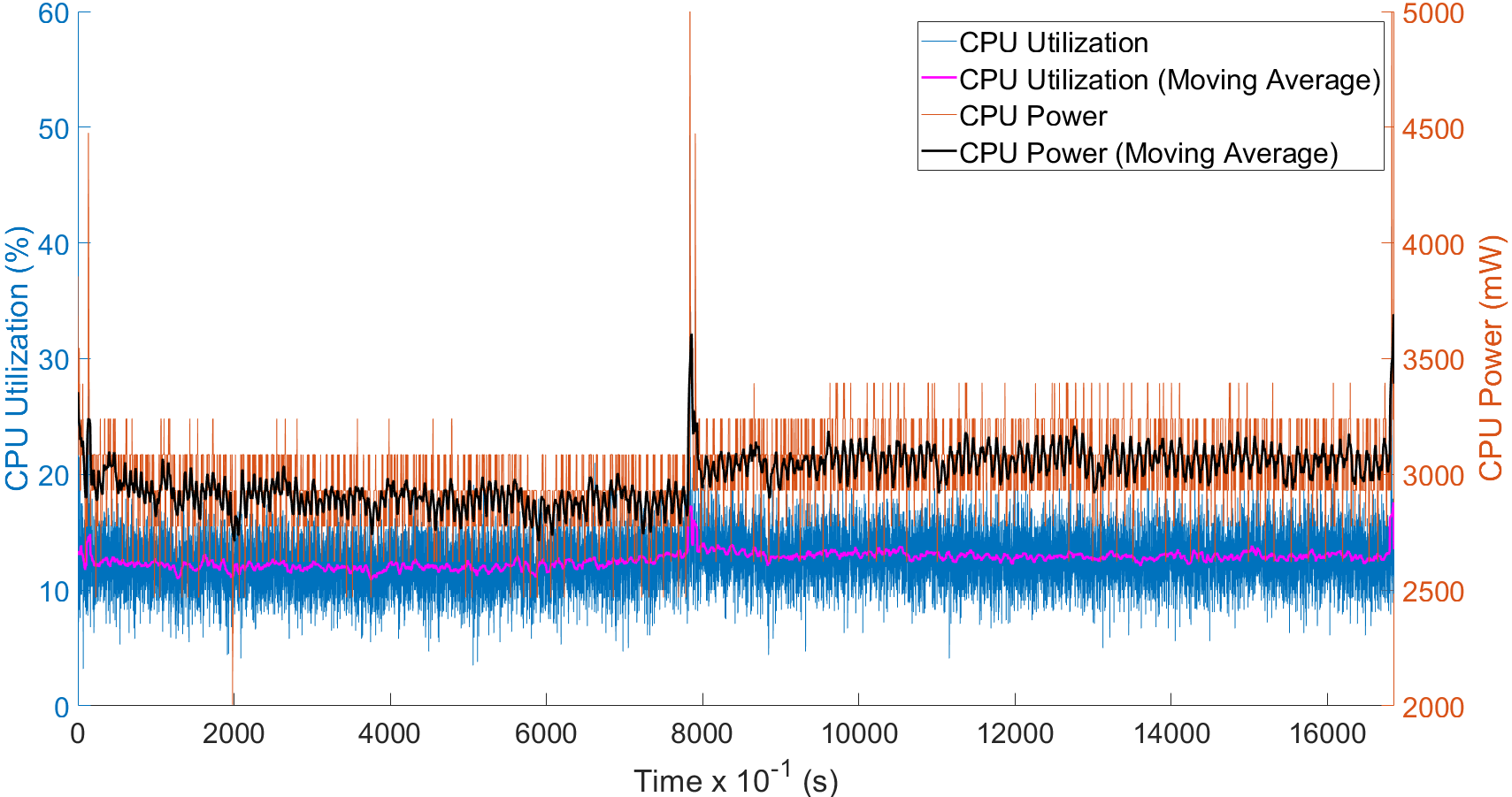} 
    \\[\abovecaptionskip]
    \small (b)
  \end{tabular}
    \caption{Computational profile data of Gmapping SLAM. (a) Execution time (User Time and System Times) for an activation period of 0.1 sec and corresponding RAM utilization. (b) CPU Utilization and Power measurements.}
  \label{fig:gmap}
\end{figure}

Moving to the Gmapping SLAM kernel, which also uses our real-world data, it is evident from the profile in Figure \ref{fig:gmap} that the variation is far less when compared to LOAM and it has a lower utilization and execution time for the Xavier AGX platform. However, due to the powerful computational capabilities of Zotac, we don't see this difference between LOAM and Gmapping in Zotac. The RAM usage of this kernel is also much more stable. However, the quality of maps produced by Gmapping is significantly lower than LOAM, which is a critical trade off. Indeed, in the authors' experience the maps produced by Gmapping are often insufficient for robot navigation, especially in large and/or complex environments. The summary of processes and threads of Gmapping in Table \ref{thread_gmap}, (same interpretation as Table \ref{thread_loam}) also conveys the relative simplicity of Gmapping, being mostly a single process kernel as the other processes are just static transform broadcasters and contribute very little to the computational load. For the other two platforms, the behaviour and utilization of Gmapping SLAM is very similar and therefore we omit the detailed distribution plots for brevity. 
A fundamental insight is that the Gmapping is a much simplistic version of SLAM and also 2D. Therefore it is expected that it has very small variation across different platforms but prior to this study we have not found any findings that proves it with real world dataset. This finding is quite useful for users when planning for deployment.

\begin{table}
\caption{Process and Thread's profiling summary of SLAM Gmapping in Xavier AGX platform}
\label{thread_gmap}     
\resizebox{\columnwidth}{!}{%
\begin{tabular}{lllll}
\hline\noalign{\smallskip}
 Process & Gmapping  & Static Transform  & Base Link-Laser   & Odom Map   \\
     & main node & Publisher & Transform & Broadcaster \\
\noalign{\smallskip}\hline\noalign{\smallskip}
 Number of Threads & 4 & 4 & 4 & 4 \\ 
  Number of Threads & 8 & 4 & 4 & 4 \\ 
  \% Util. of dominant thread & 32.82 & 87.87 & 87.96 & 88.27 \\
  \% of Execution Time & 97.5 & 0.8 & 0.8 & 0.7 \\
  Avg. Execution Time & 0.10231 & 0.00156 & 0.0014 & 0.00094\\ 
  Variance in Exec Time & 0.00385 & 0.1.5E-5 & 1.4E-5 & 9.5E-6 \\
\noalign{\smallskip}\hline
\end{tabular}}
\end{table}

\begin{figure}
 \centerline{\includegraphics[width=3.6in]{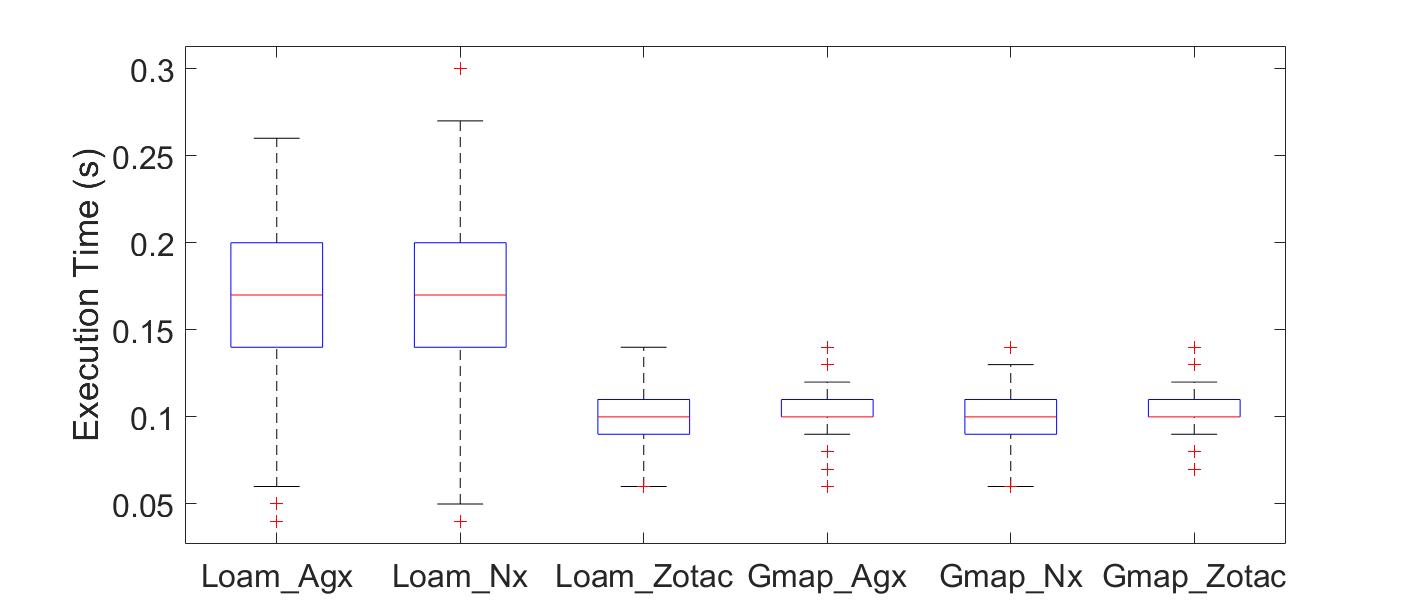}}
 \caption{Boxplots for execution times for SLAM kernels for different hardware platforms. The x-axis denotes the obstacles density that is the density of obstacles per sqm}
 \label{slam_box}
\end{figure}

To conclude, Figure~\ref{slam_box} presents the comparison of execution times for the two SLAM kernels over our three computational platforms.

\subsubsection{RRT-based Path Planning}
\begin{figure*}[tb]
\centering
\includegraphics[width=0.8\textwidth]{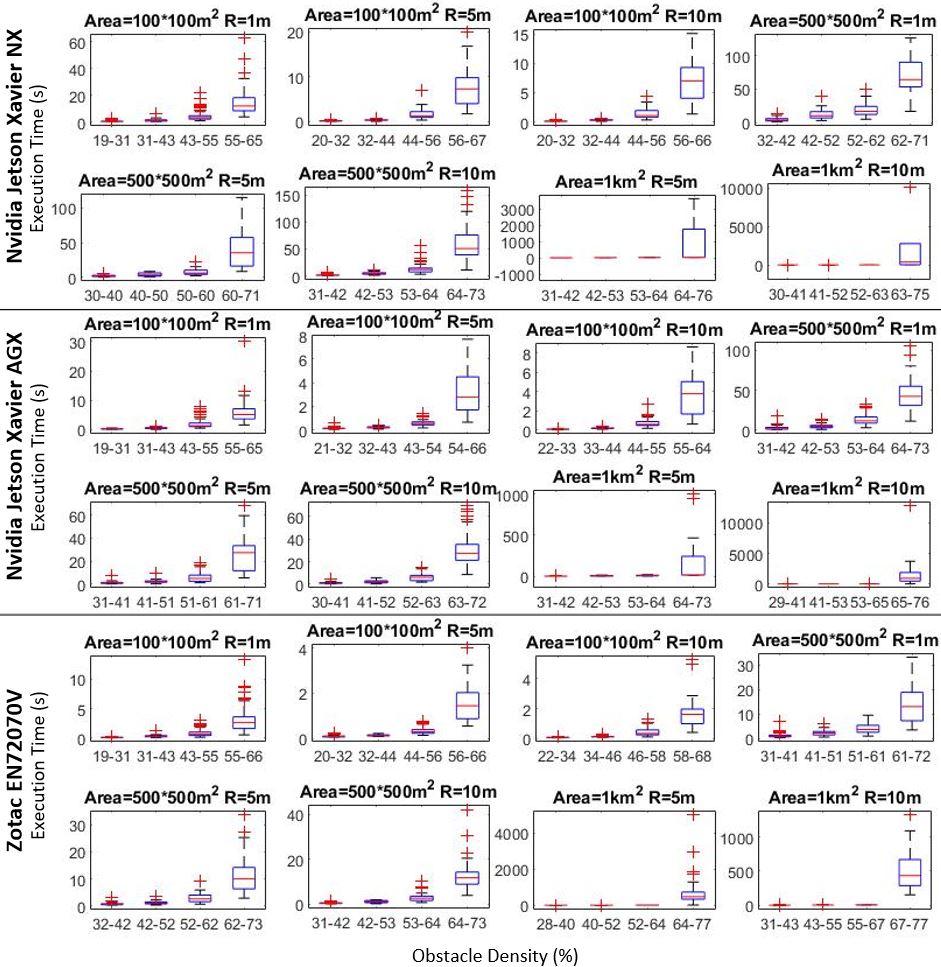}
\caption{Boxplots of execution or cpu user times of RRT path planning experiments}
 \label{rrt_box}
\end{figure*}

The RRT path planner was executed as a standalone python3 code using the libraries from \cite{rrtalgo}. As a standalone python code, only a single thread was monitored in all the platforms. For an RRT planner, as specified in Section III, step/sampling size (R), environment size, and obstacle density were varied for each of the computational platforms. The resulting computational profiles in Figure \ref{rrt_box} show the effect of these parameters on the execution times. Typically, the sampling or step size must be determined arbitrarily based on prior experience in related use cases. Generally, a lower sampling size will result in higher execution times but a larger sampling size can also generate infeasible paths as small environment features (such as obstacles) can be overlooked. Moreover, as the density of obstacles and size of environment increase, the execution times also increase. A lower sampling size in a large area, such as a square kilometer, is not desirable to find paths in a timely manner. However, with the obstacle density increasing, higher sampling sizes also tend to generate infeasible paths. Thus, knowledge of the average and worst case execution times in such scenarios are a must to plan paths in a timely manner for autonomous robots in complex and/or dynamic environments. Obstacle Density is a crucial environmental parameter for path planning and its relation to response times are rarely studied in literature.

\subsubsection{Task Allocation}
\begin{figure*}[tb]
\centering
\includegraphics[width=0.99\textwidth, height=300pt]{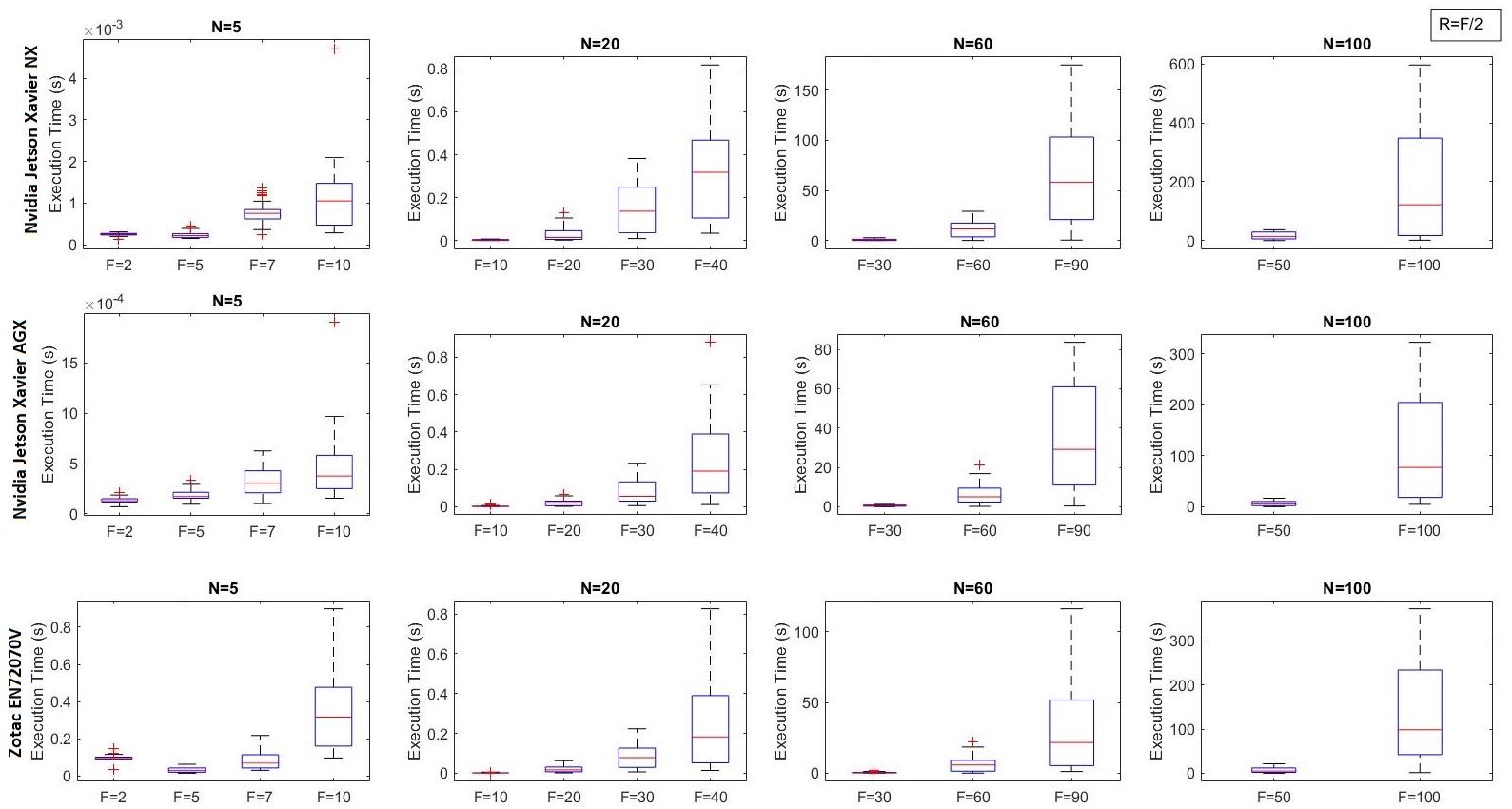}
\caption{Boxplots for execution or cpu user times of multi robot task allocation experiments}
 \label{ta_box}
\end{figure*}

\begin{figure}
 \centerline{\includegraphics[width=2.5in]{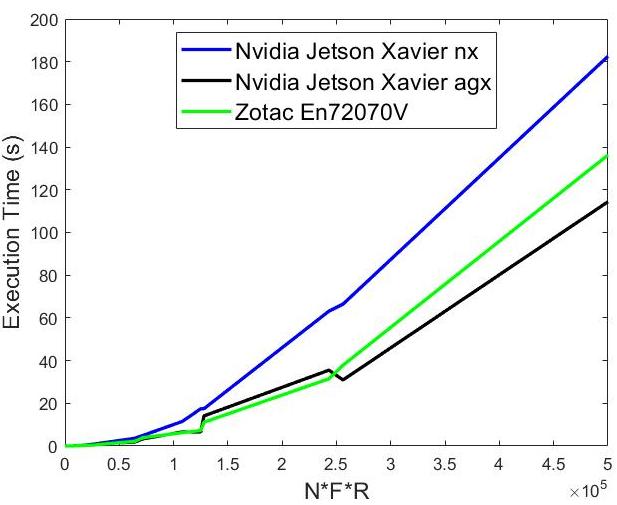}}
 \caption{Complexity of task Allocation kernels}
 \label{ta_comp}
\end{figure}

The task allocation kernel is written in C code by the authors following the algorithms from the task allocation problem in \cite{Williams2017}. As a standalone C code, this kernel was also monitored as a single thread in all the platforms. As stated in Section III, we vary the robot cardinality, $N$, the requirement cardinality $R$, and functionality cardinality $F$ to generate a set of task allocation problems with varying difficulty.  Specifically, our test cases are built by setting $N \times average(F) = F$ and $R = F/2 $ . We choose N from 5 to 100. Figure \ref{ta_box} depicts the execution time variations due to these parameter changes. In addition, Figure~\ref{ta_comp} validates the polynomial complexity of the kernel, originally proved in \cite{Williams2017}.
The useful insight for a single threaded kernel is quite straightforward from this data. The More power and frequency for each cores the platform has, the faster is completes the tasks. That's why we see the execution times get lower from Nx to AGX to Zotac. 

\begin{figure}
 \centerline{\includegraphics[width=5in]{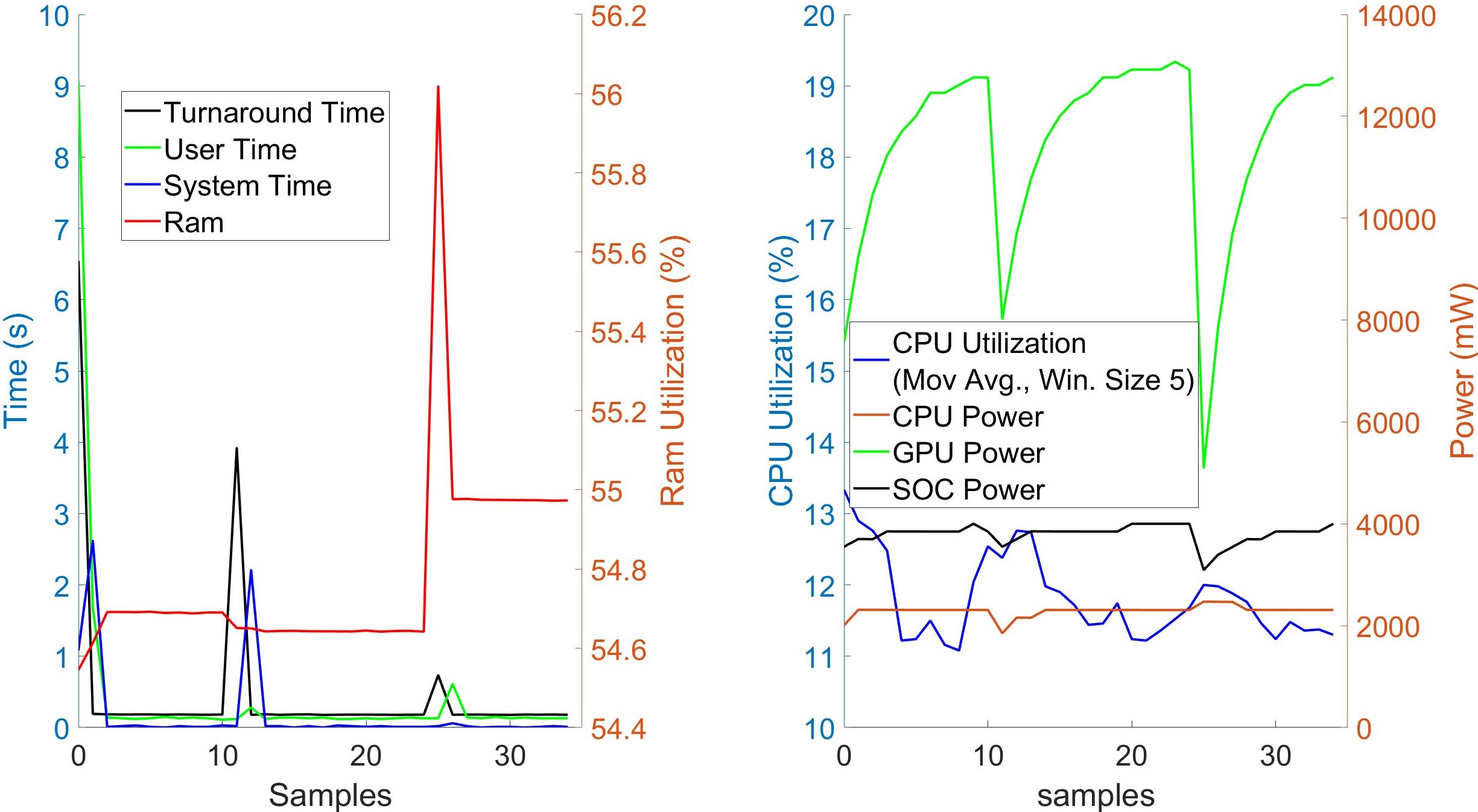}}
 \caption{Midas depth estimate kernel profile in Xavier AGX. (left) Turnaround, Execution (User and System) and RAM utilization of Midas. (right) CPU utilization and CPU, GPU, SoC power consumption profiles.}
 \label{fig:midas}
\end{figure}

\begin{figure}
 \centerline{\includegraphics[width=5in]{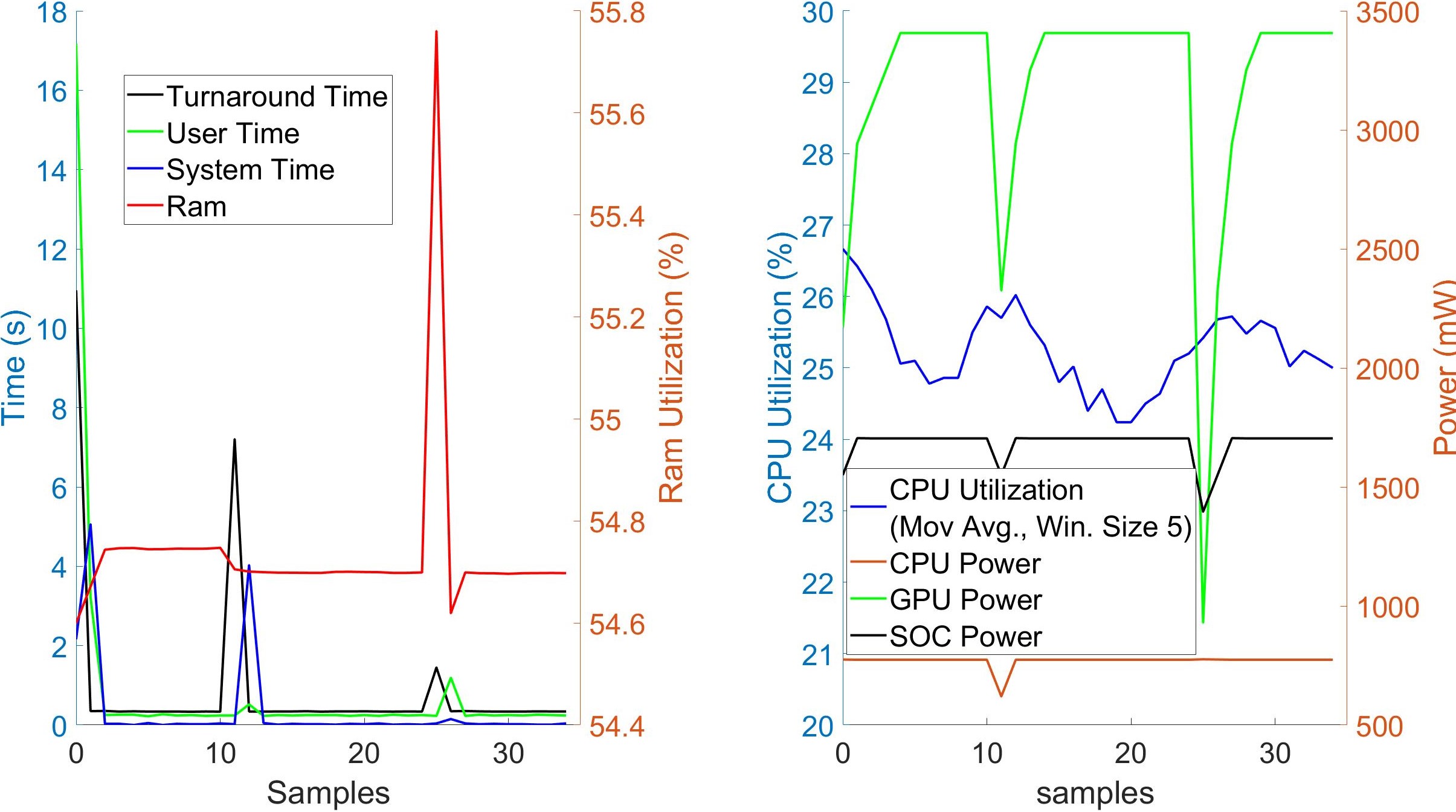}}
 \caption{Midas depth estimate kernel profile at low (15W) power mode of Xavier AGX. (left) Turnaround, Execution (User and System) and RAM utilization of Midas. (right) CPU utilization and CPU, GPU, SoC power consumption profiles.}
 \label{fig:midas10}
\end{figure}

\subsubsection{Optical Flow and Monocular Depth}

The CV kernels were not operated with a particular activation period but on a typical while loop with no timing constraints. Therefore, we also provide the turnaround time information for these kernels. Both kernels are also run standalone utilizing libraries from the Github repositories \cite{Ranftl2019,sun2018} and take as input vision data from our real-world dataset.  Figures \ref{fig:midas} and \ref{fig:midas10} show the computation and power profiles for the monocular depth estimation kernel in maximum and low power modes respectively. The first insight is the high usage of the shared CPU/GPU memory on the Xavier platform, which is highly utilized and almost completely consumed for Pwc-Net. The high utilization of RAM is due to fact that memory is shared between the CPU and GPU via a SoC in the Xavier platforms and these kernels are using GPU intensive PyTorch libraries.  Also there is a significant increase in the execution times and utilization of CPU in low power mode. The interesting aspect from Figure \ref{fig:midas} and \ref{fig:midas10} is that there are spikes of CPU usage in both of the power modes.

This behavior is caused by the insertion of two out of context images in the dataset of our environment. The regular images that represent the steady parts of the graphs are from our own collected data in which the environment is gradually changing as the robot moves. However, when it faces an out-of-context or new environmental feature set, it spends a lot of time processing the change in the CPU. There is also evidence of a CPU bottleneck issue after analyzing this portion of execution in the NVIDIA Nsight system and thus there is an opportunity of optimizing the kernel for usage on embedded heterogeneous platforms like Xavier.     

\begin{figure}
 \centerline{\includegraphics[width=5.0in]{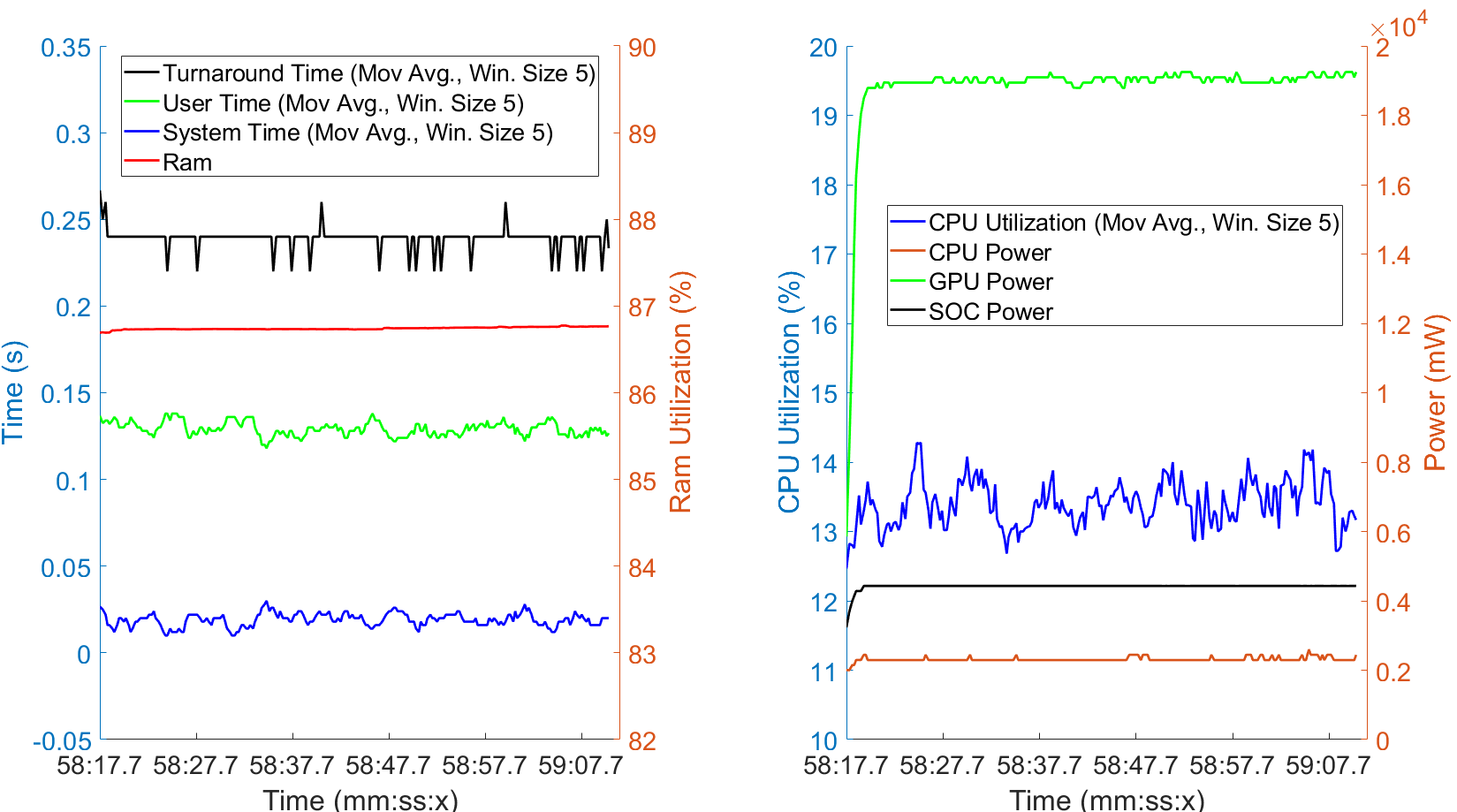}}
 \caption{Pwc-net optical flow estimation kernel profile in Xavier AGX. (left) Turnaround, Execution (User and System) and RAM utilization of Midas. (right) CPU utilization and CPU, GPU, SoC power consumption profiles.}
 \label{fig:pwc}
\end{figure}
\begin{figure}
 \centerline{\includegraphics[width=5.0in]{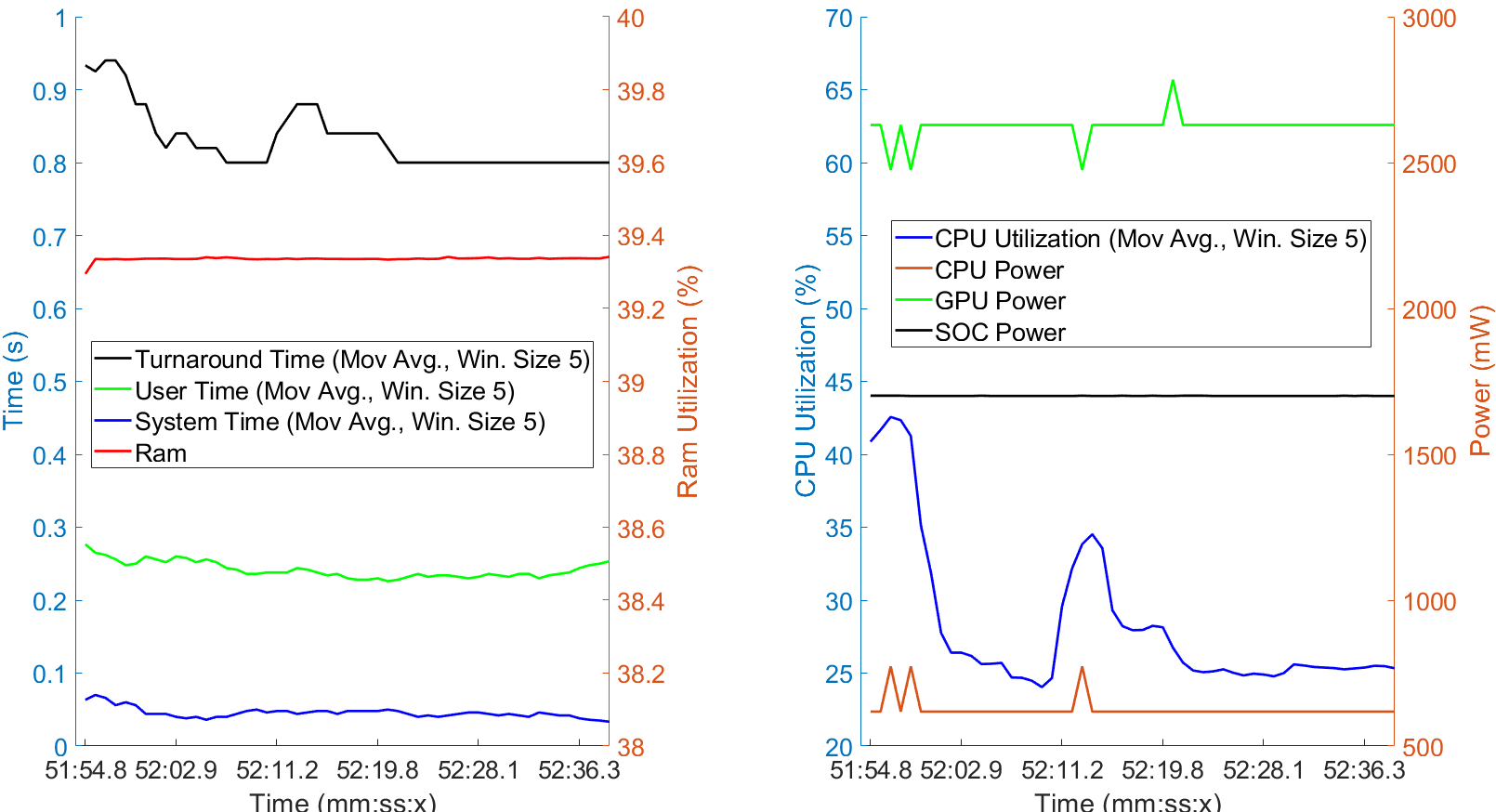}}
 \caption{Pwc-net optical flow estimation kernel profile at low (15W) power mode in Xavier AGX. (left) Turnaround, Execution (User and System) and RAM utilization of Midas. (right) CPU utilization and CPU, GPU, SoC power consumption profiles.}
 \label{fig:pwc10}
\end{figure}

The optical flow estimation kernel is conducted in a similar manner as the depth estimation kernel without the insertion of out-of-context images. It is fed with a set of sequential images from the same path the SLAM kernels were conducted on. The profiling information is visualized in Figure \ref{fig:pwc} and \ref{fig:pwc10}. Here, there is a slight variation in computational loads, which is attributed to changing scenarios in the dataset. There is a shift to CPU computation from GPU during those instances, which contributes to the increase in CPU execution time which is more evident in the low power mode. Moreover, the GPU resources are also utilized at a minimum level which is evident from the utilization summary of several GPU parameters. The achieved Occupancy for MiDaS and Pwc-Net are respectively 25.28\% and 20.77\% and all the functional unit utilization are low or idle (0-2) when analyzed using the Nvidia Profiler. The Warp Execution Efficiency for MiDaS and Pwc-Net were observed to be 96 and 95\% whereas the Shared Utilization were low or mid level (0-6) for both cases. These data were observed for 61 and 35 kernel threads for MiDaS and Pwc-Net respectively and given as a summary or average here. For the Zotac platform, where we have to measure the GPU parameters using \emph{Nvidia-smi} interface, we observe low utilization of the GPU resources. For Midas in Zotac, the GPU utilization is 35.5 with GPU memory utilization at 15.35775 \% and a much high GPU power usage at 72 W. For Pwc-Net in Zotac, the GPU utilization is 16.4 \%, memory utilization at 16.9 \%, and GPU power usage at 46 W. 

\begin{figure}
 \centering
  \begin{tabular}{@{}c@{}}
    \includegraphics[width=.7\linewidth]{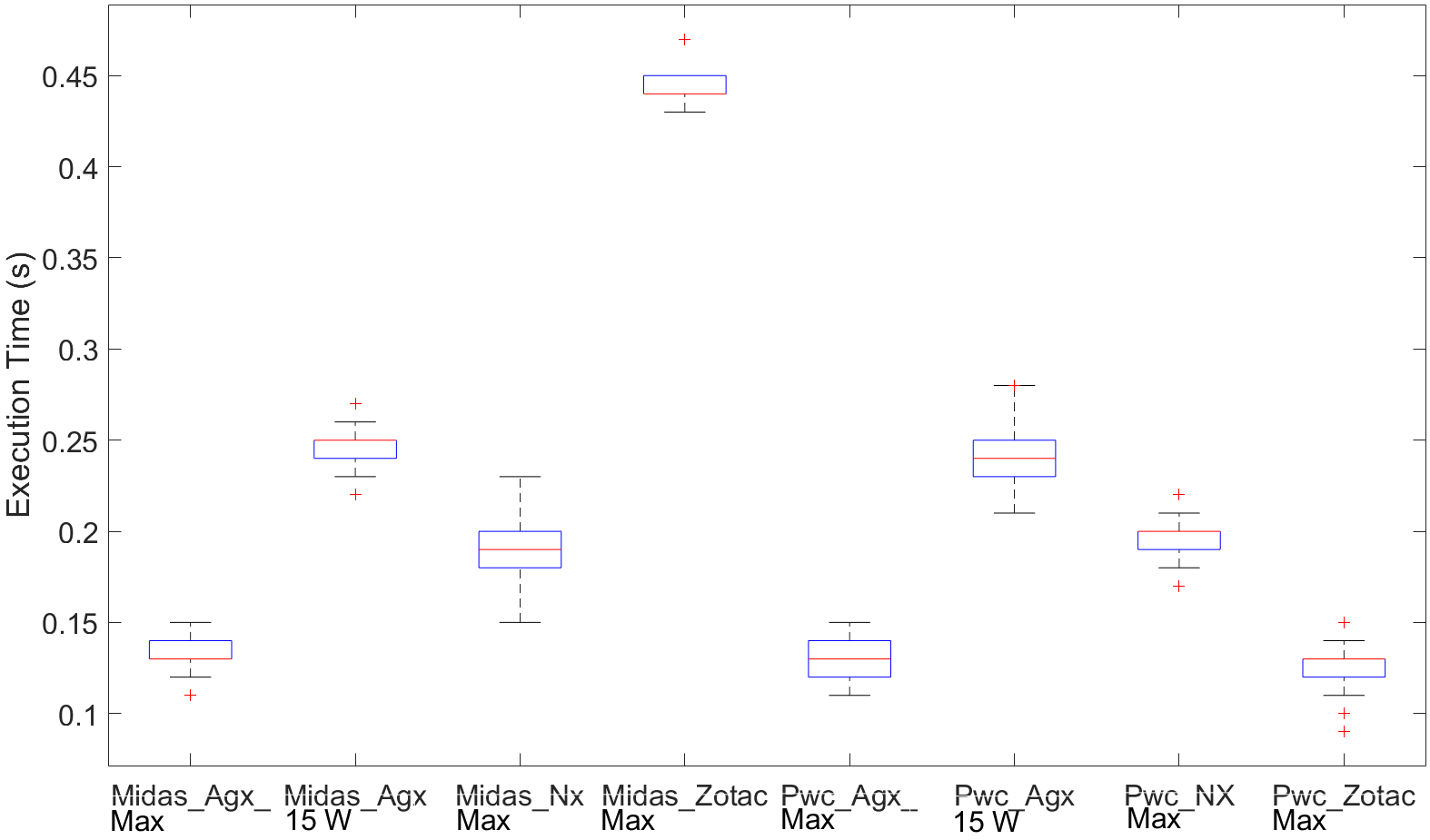} 
    \\[\abovecaptionskip]
    \small (a)
  \end{tabular}
  \begin{tabular}{@{}c@{}}
    \includegraphics[width=.7\linewidth]{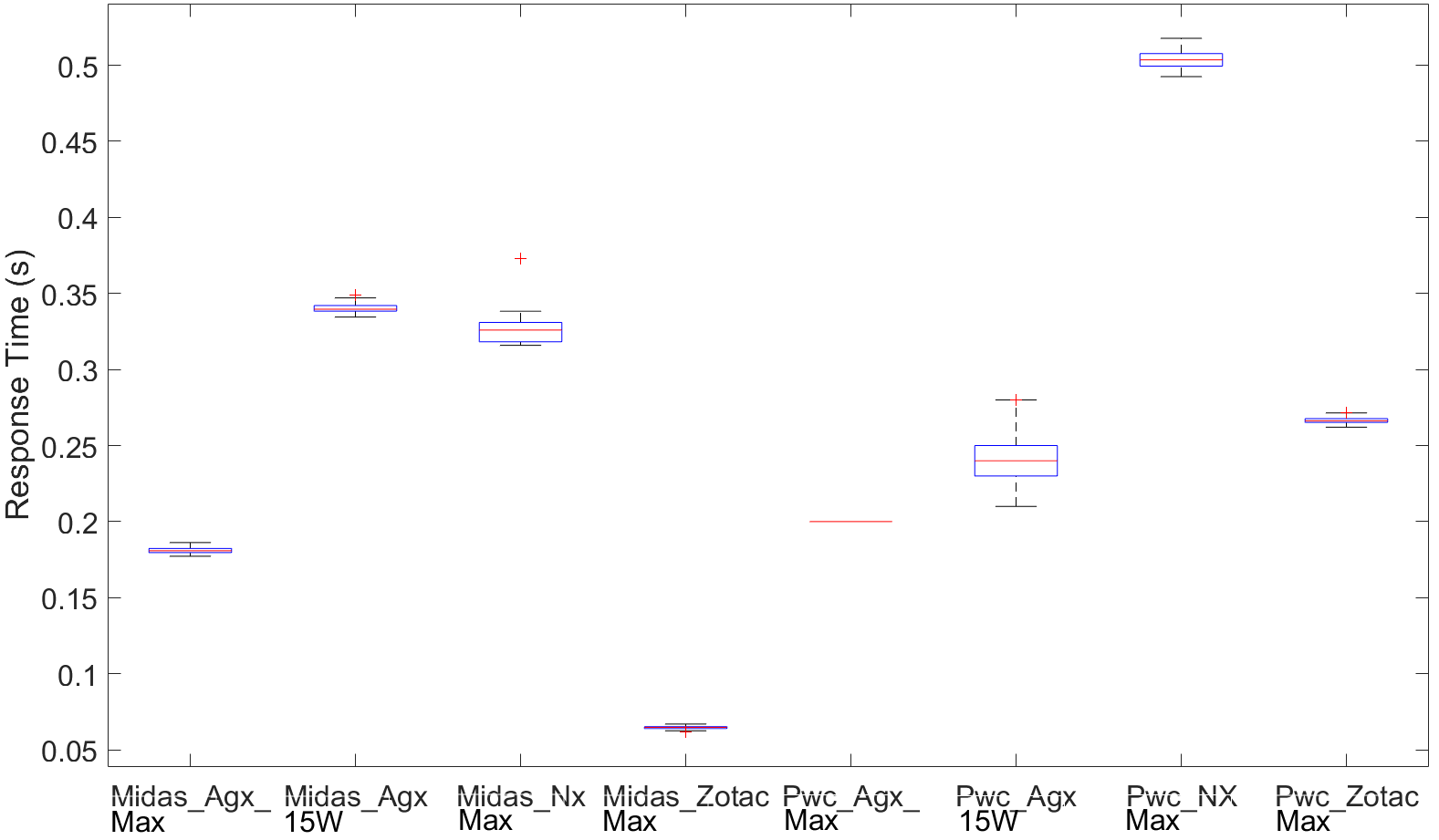} 
    \\[\abovecaptionskip]
    \small (b)
  \end{tabular}
    \caption{(a) Execution times. (b) Response/Turnaround times for vision kernels. Max=maximum power.}
  \label{vis_box}
\end{figure}

We depict the distribution of the different parameters over the dataset of both of these CV kernels in Figures \ref{fig:midas}, \ref{fig:midas10}, \ref{fig:pwc} and \ref{fig:pwc10} for the Xavier AGX platform only. For the other platforms, we observe similar spikes in computation for the out-of-context images for depth estimation. For the optical flow estimation, the execution times are also very similar. However, the resource utilization for the Zotac is quite low compared to the Xaviers and the data as well and cause of this is discussed in the next section. We again omit the similar figures of distribution of execution time and resource utilization for Xavier NX and Zotac for brevity.  Finally we present in Figure \ref{vis_box} the boxplots for the computer vision kernels across our computational platforms to represent these comparisons. 

It is visible that in the Zotac platform, the execution times are higher whereas the overall response times are proportionally lower. This is attributed to the fact that the higher CPU core count of Zotac yields more parallelism of the vision kernels in Zotac compared to the Xaviers and thus overcomes the CPU bottleneck issue in the Xaviers to some extent. 

\begin{remark}
The data collection experimentation and also corresponding images related to SLAM and MiDaS are depicted in the video https://youtu.be/LvapXWqEAqU. This gives a clearer picture of the environment the experiments were conducted in, including both outdoor and indoor scenarios.
\end{remark}

\section{Discussions} \label{sec:disc}

\begin{figure}
 \centerline{\includegraphics[width=6.0in]{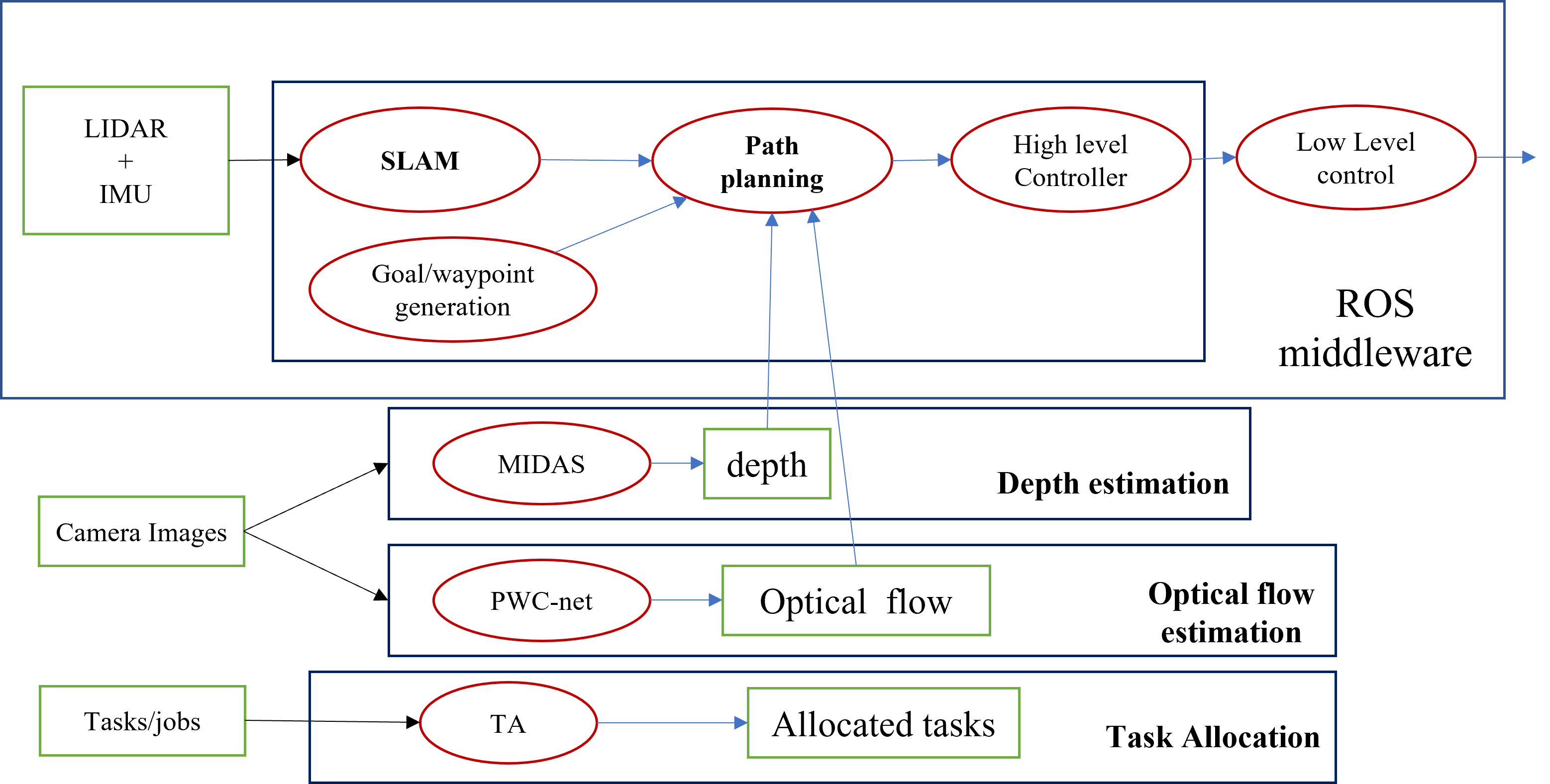}}
 \caption{An example DAG comprised of the computational kernels presented in this paper.}
 \label{dag-exp}
\end{figure}

Our profiling efforts yield several significant insights that we argue are fundamental to our goal of guaranteeing the timeliness and correctness of complex autonomous behaviors in robotic systems. First, from all the data presented earlier as well as the summary in Table \ref{summary} it is quite evident that the operation of these kernels simultaneously, which is \emph{necessary} for complex robotic applications, will degrade performance and yield missed deadlines (which are equal to the activation periods), especially in low power scenarios. While the overall utilization is not merely a summation of the individual utilizations, the surprising changes in response times in certain scenario can generate these deadline misses.  A change in activation periods and appropriate scheduling as well as prior knowledge of these changes in response/execution times will be necessary to operate a combination of vision and CPU-bound kernels, which if done naively may yield catastrophic impacts on performance and safety of autonomous behaviors.  This motivates our ongoing efforts to develop a \emph{dynamic} scheduling parameter adjustment for autonomous robots based on the data presented in this work. Fig \ref{dag-exp} refers to an example application using the computational kernels that can be scheduled in a Directed Acyclic Graph (DAG) scheduling framework such as implemented in ROSCH \cite{saito2019}. In any such DAG scheduler, the comprehensive knowledge of individual execution times are a must for designing a dynamic computationally aware scheduler \cite{zhao2020,stefano2020}.

Next, the LOAM kernel shows a spatial correlation in the event of loop closure and revisiting environmental cues with a drop of the execution time and other related parameters. However, the memory usage continues growing with the traversal of the environment. In about 30 minutes of traversal, the RAM usage grows from 6\% to 12\% in the Xavier AGX platform, with similar growth in Xavier NX and Zotac. On the other hand, the Gmapping SLAM kernel shows a quite stable behavior, but with significantly lower map quality. Interestingly, in low power mode the LOAM kernel is inoperable at the default activation period of 0.1 seconds as 100\% utilization is seen, resulting in missed execution deadlines.  These SLAM insights allow for unique opportunities such as dynamic scheduling parameter adjustments that anticipate loop closure events, or robot mission planners that balance location revisitation with the availability of computational resources. 

Shifting to the task allocation and RRT path planning kernels, we see an even greater variation in both average and WCET as evident from the boxplots in Figures~\ref{rrt_box} and \ref{ta_box}, with these effects being directly dependent on a robot's operational environment and/or objectives.  Once again such insights open up dynamic resource management opportunities, such as controlling the quality of planned paths based on computational resources and safety margins, or mission planners that anticipate task loads in an environment and act to mitigate impacts on kernel scheduling. The single threaded processes such as RRT and TA can have differences in execution times between the AGX and NX platforms because the NX is more conservative in power consumption than the AGX. This can affect the execution of kernels specially for higher periods of executions.  That is evident in Fig. \ref{rrt_box}. Moreover, we can utilize the GPU for optimization of RRT and TA by running the critical sections on the GPU.

Focusing now on the CV kernels, significant variation in resource utilization is observed with changing Xavier power modes. For example, the execution time increases 3 fold both in the Midas depth estimation and optical flow estimation.  The vision kernels also show a classic example of GPU starvation and CPU bottleneck in Xavier hardware. Parts of the algorithm are sequential with no parallel operation implemented, in particular, portions dedicated to preprocessing the images before calling the PyTorch libraries. Most of the GPU resources are underutilized while occupying a huge amount of shared memory space. This usage of a huge amount of memory calls for a conflict resolution in CPU and GPU memory access, which is investigated in very recent literature \cite{Fang2020},  \cite{6237036}, \cite{6714422}, and can be put into action using the data from this paper. In the Zotac system, which has a discrete GPU unlike the Xaviers, the CV kernels show much better performance and lower CPU bottleneck issue, however the Zotac platform is not viable for aerial robots.  This final point indicates that for vision kernels there may be a strong case for offloading computation to more capable agents (e.g., from an aerial robot to a ground robot or the cloud).
Although it bridges the gap between theoretical and practical  usage of robotic computational kernels, some of the results in this paper are quite predictable. However, the Pwc-Net results in fig. \ref{vis_box} depicts quite the unpredictable behaviour. It reflects that only using powerful hardware with discrete GPU cannot yield better response times for all the pytorch enabled computation and some DNN implementations are more compatible to the embedded hardware like Xaviers than the others. It however uses much less RAM in Zotac due to its discrete GPU memory.

A final insight from the summary in Table \ref{summary} is that the memory conflict for concurrent operation of the kernels may also influence the schedulability and predictability of operation similar to the typical CPU resource management schemes.

\begin{remark}
An important note is the presence of noise in our data, stemming primarily from our data sampling mechanism and file writing operations for profiling result collection. The use of an average differential mechanism keeps the variance of that noise present in data. However, we have also profiled the noise and account for its average value. Specifically, we can quantify that the error margin of our presented data is 2.36\% for CPU utilization, 0.015 \% for RAM utilization and 6.9\% for CPU power. The execution times are devoid of this error.
\end{remark}

\begin{table}
\caption{Summary of Benchmarks and the Profiling Results}
\label{summary}
\centering
\begin{tabular}{lllll}
\hline\noalign{\smallskip}
  Platform  & Kernels & Power    & CPU & RAM    \\
    & &      Mode    &   Util  (\%)   & Util(\%) \\
\noalign{\smallskip}\hline\noalign{\smallskip}
 & LOAM SLAM & Max  & 32 & 5-20\\
Nvidia &   Gmapping SLAM & Max & 22 & 12 \\
  Xavier & RRT Path planning & Max & 13.2 & 17.1 \\
  AGX & Task Allocation & Max & 14 & 17.7 \\
 & MiDaS depth est. & Max  & 12.2 & 56 \\
     &   &  15W   & 26  & 55 \\
  & Pwc-Net Optical Flow est. & Max  & 13.4 & 86.7 \\
    &   & 15W   &  27.8 & 40 \\
\noalign{\smallskip}\hline\noalign{\smallskip}
 & LOAM SLAM & Max  & 57 & 10-22\\
Nvidia &   Gmapping SLAM & Max & 31 & 32 \\
  Xavier & RRT Path planning & Max & 18 & 24.5 \\
  NX & Task Allocation & Max & 20 & 23.5 \\
 & MiDaS depth est. & Max  & 10.2 & 91 \\
  & Pwc-Net Optical Flow est. & Max  & 19 & 89.4 \\
\noalign{\smallskip}\hline\noalign{\smallskip}
  & LOAM SLAM & Max  & 18.7 & 6-11\\
 &   Gmapping SLAM & Max & 15.3 & 8 \\
  Zotac & RRT Path planning & Max & 8 & 4.5 \\
   & Task Allocation & Max & 8.3 & 8.4 \\
 & MiDaS depth est. & Max  & 27.2 & 12.1 \\
  & Pwc-Net Optical Flow est. & Max  & 6.9 & 13.5 \\
\noalign{\smallskip}\hline
\end{tabular}
\end{table}
Finally we also derive additional insights into scheduling, optimization and resource management for future deployment of these kernels simultaneously. 

a) GPU-intensive kernels: The DNN based algorithms have various ways of optimization depending on the hardware such as utilizing the vision accelerators in the Xaviers as well as generalized approximation techniques such as matrix factorization, compression. Moreover, GPU intensive kernels also have environmental correlation which can be modeled and further studied with empirical data such as the other kernels in this paper.

b) CPU-intensive kernels:The analysis in this paper utilizes the default CFS scheduler of the Linux kernel. Specifically, the parameters which could be dynamically adjusted for scheduling these computational kernels via Linux schedulers are priorities, scheduler tick, vruntime, deadlines and the scheduling mechanism itself. Therefore, our followup work is on analyzing real time schedulers such as FIFO/RR/EDF and in that case we will also define hyperperiod of tasksets and target latencies and also analyse the effects of these schedulers on the same para,meters as this paper but with a combinatorial implementation of the computational kernels. Such experiments will generate a  learned computational map which can be used for our ultimate objective of computation aware autonomy, as specified in Fig \ref{fig:framework} There are several scheduling algorithms that can be utilized such as deadline based hard real time, soft real time and novel environment aware schedulers. Usage of RT-patches and RTOS, CPU isolation, memory locking are also some of the possible options. The applicability of these will depend heavily on the mission objectives. 

c) Combining a and b: Moreover, the Jetson Xavier architecture shares the RAM between the CPU and GPU, therefore the GPU intensive kernels have a much higher RAM utilization. This could pose an issue when used combinatorially with other kernels and some literature which address these memory access issues in heterogeneous architectures are cited in the discussions[23-25].

\section{Conclusion}
Autonomous systems cannot treat computation as a black box anymore as has been done in most of the systems deployed in present time. Our study provides a detailed preliminary insight into the means of achieving the timeliness guarantee.
This paper aims to gain deeper insights into the robotics computational kernels in real environments. To limit the experiments’ extent to meaningful and interpretable content, we choose a set of representative computational kernels needed for broader autonomy. Most of them are from existing libraries except the Task Allocation kernel (which was developed by the authors due to unavailability of C code implementation). We rigorously analyze the effects of environmental parameters such as depth and motion features, complexities of indoor/outdoor, size and obstacle densities. While the results were mostly not surprising, our quantified profiles fill the gap of current literature for timely and safe autonomy. The work on environmental correlation of SLAM and RRT path planning kernels are of vital importance for practical applications. Moreover, for the application of a DAG scheduling comprised of these computational kernels, this individual execution times measurements are considered as input data for he schedulers.
We also use existing and well renowned tools such as psutil so that the methodologies are not questionable.  We have not changed the many different features on the hardware platforms, but still from the results, it can be deduced that the number of cores, maximum power consumption, operating frequency and also architecture difference such as integrated or discrete GPU all have an effect on the performance. 

\bibliographystyle{spphys}
\bibliography{arxiv}
\end{document}